\documentclass[10pt]{article} 
\usepackage[preprint]{tmlr}

\usepackage{times}
\usepackage{latexsym}
\usepackage[T1]{fontenc}
\usepackage[utf8]{inputenc}
\usepackage{microtype}
\usepackage{inconsolata}
\usepackage{graphicx}
\usepackage{booktabs}
\usepackage{hyperref}
\usepackage{amssymb}
\usepackage{amsmath} 
\usepackage{tikz}
\usepackage[edges]{forest} 
\usepackage{xcolor}
\usepackage{adjustbox}
\usepackage{array}
\usepackage{url}

\newcommand{\itembreak}{\\[4pt]}
\definecolor{colRoot}{RGB}{40, 60, 100}

\definecolor{colLife}{RGB}{41, 128, 185}
\definecolor{bgLife}{RGB}{235, 245, 250}

\definecolor{colTax}{RGB}{142, 68, 173}
\definecolor{bgTax}{RGB}{245, 235, 250}

\definecolor{colEco}{RGB}{39, 174, 96}
\definecolor{bgEco}{RGB}{235, 250, 240}

\definecolor{colChal}{RGB}{211, 84, 0}
\definecolor{bgChal}{RGB}{253, 240, 230}

\usepackage{amsmath,amsfonts,bm}

\def\eqref#1{equation~\ref{#1}}

\def\1{\bm{1}}

\DeclareMathAlphabet{\mathsfit}{\encodingdefault}{\sfdefault}{m}{sl}
\SetMathAlphabet{\mathsfit}{bold}{\encodingdefault}{\sfdefault}{bx}{n}

\title{A Systematic Survey of Agentic Skills: Architecture, Lifecycle, and Security}

\author{\name Sanket Badhe \email sanketbadhe@google.com \\
      \addr Google LLC
      \AND
      \name Deep Shah \email shahdeep@google.com \\
      \addr Google LLC
      \AND
      \name Priyanka Tiwari \email tiwarip@alumni.purdue.edu \\
      \addr Purdue University
      \AND
      \name Nehal Kathrotia \email nehalk@google.com \\
      \addr Google LLC}

\def\month{08}  
\def\year{2026} 
\def\openreview{\url{https://openreview.net/forum?id=XXXX}}

\begin{document}

\maketitle

\begin{abstract}
Autonomous large language model (LLM) agents increasingly face reliability, context consumption, and execution stability bottlenecks when deployed on complex, long-horizon tasks. While monolithic prompt engineering and stateless tool-calling paradigms struggle to scale, the field is rapidly converging toward \emph{agentic skills}: modular procedural abstractions that externalize execution knowledge into reusable, executable, and portable artifacts. This paper establishes a unified systems foundation and reference architecture for the agentic skills ecosystem. We formalize skills as externalized procedural knowledge bridging high-level cognitive planning with deterministic execution environments, and systematically delineate the architecture across a nine-stage lifecycle: autonomous discovery, authoring and representation formats, memory storage, dynamic retrieval and routing, composition and orchestration, execution and repair, lifelong adaptation, empirical evaluation, and security governance. We further examine marketplace dynamics, public registries, and emerging adversarial threat vectors, alongside runtime verification and defense mechanisms. Finally, we categorize system implementations across software engineering, operating system navigation, embodied robotics, and scientific discovery, while highlighting critical open challenges in continual learning and benchmark realism. This work establishes agentic skills as a foundational paradigm for building scalable, robust, and verifiable autonomous language agents.
\end{abstract}

\section{Introduction}

\subsection{Motivation: The Reliability and Reusability Bottlenecks in Language Agents}
Large language model (LLM) agents, stateful systems that pursue goals over many steps by interleaving reasoning with tool calls, have become a dominant approach to automating open-ended workflows \citep{luo2025largelanguagemodelagent}. Their performance degrades sharply as task horizons lengthen. Failure traces show that losses are unevenly distributed across steps: agents are disproportionately vulnerable to early-stage execution and path-resolution errors that propagate downstream and contaminate subsequent reasoning \citep{longhorizonmirage2026, liu2026agenticskillsworkwild}. Because agents rarely detect that an early step was wrong, they enter a no-recovery regime, looping or exhausting recursion limits rather than revisiting the faulty premise \citep{norecoverybottleneck2026}.

A second bottleneck compounds the first: nothing accumulates. An agent that completes a task successfully gains no durable advantage the next time it encounters the same task, because procedural knowledge remains locked in transcripts discarded at session boundaries. Each episode re-derives routines the agent has already executed correctly, an observation motivating the recent wave of lifelong-learning agent frameworks \citep{yang2026autoskill, reme2025, continuallearningmemory2026}. Neither standard remedy closes the gap. Placing procedures in the context window imposes a token cost that grows with the instruction set and provides no mechanism for loading only what a task requires \citep{song2024easytool, skillreducer2026}. Fine-tuning writes procedural knowledge into weights, whereas the routines agents need are tied to tooling, APIs, and organizational conventions that change faster than retraining cycles accommodate \citep{zhuang2026agentclickskillbasedhumanintheloopreview}. 

Externalized executable skills target this second bottleneck. By packaging activation conditions, procedural instructions, tool declarations, and executable assets into modular artifacts stored outside the model, skills separate high-level planning from deterministic execution \citep{liu2024skillact}, persist across sessions, load only when relevant, and can be revised without retraining. This paper establishes a systems framework for how that abstraction is formalized, authored, retrieved, executed, and secured.

\subsection{What Is an Agentic Skill? Formalizing the Abstraction}
\label{sec:intro_abstractions}

To establish a rigorous theoretical foundation across the literature, we formalize an externalized agentic skill as a six-tuple:
\begin{equation}
s = (\mathcal{A}, \mathcal{I}, \mathcal{C}, \mathcal{T}, \pi, \mathcal{E})
\end{equation}
where $\mathcal{A}$ denotes the activation condition, $\mathcal{I}$ represents procedural instructions or guidance, $\mathcal{C}$ denotes applicability constraints and environment preconditions, $\mathcal{T}$ represents accessible tools and interfaces, $\pi$ represents the execution policy or procedural control flow, and $\mathcal{E}$ denotes the intended state transitions or post-condition effects \citep{song2024easytool, taskbench2024}.

\paragraph{The Triad Membership Criteria:}
To distinguish agentic skills from adjacent computational abstractions in language agent architectures, an artifact must satisfy three core systems-level conditions:
\begin{enumerate}
    \item \textbf{Encapsulated Modularity:} A skill is a discrete, externalized asset that bundles procedural logic, interface declarations $\mathcal{T}$, and constraints $\mathcal{C}$ into an independent package. It can be authored, versioned, validated, and updated without altering foundation model weights or modifying global agent harness code.
    \item \textbf{Late-Binding Dynamic Invocation:} Unlike static runtime components that reside permanently in the context prompt, a skill is externalized and loaded on-demand. Its inclusion in active memory is governed by the dynamic evaluation of its activation function $\mathcal{A}(x_t, g) \rightarrow \{0, 1\}$ over the agent's current state $x_t$ and active goal $g$.
    \item \textbf{Procedural State Transformation:} A skill encodes reusable operational knowledge, dictating how to execute a task and handle potential errors through policy $\pi$, rather than recording declarative facts or logging chronological histories.
\end{enumerate}

\paragraph{Disambiguating Borderline Cases:}
Applying these criteria resolves common ambiguities at the boundaries between skills and related concepts:
\begin{itemize}
    \item \textbf{Prompts vs. Skills:} A monolithic system prompt injected globally at session initialization, such as a general directive to format all responses in structured JSON, provides static agent parameterization; because it lacks modular encapsulation and late-binding retrieval, it is not a skill. Conversely, a task-specific standard operating procedure for refactoring React components, stored in a procedural library and injected into the prompt only when frontend source code is encountered, satisfies all three criteria and represents an Instructional Skill \citep{ling2026agent}.
    \item \textbf{Atomic Tools vs. Skills:} A standard API schema provided directly to the model, such as a basic web search endpoint, is a stateless functional primitive $\mathcal{T}$ functioning simply as an atomic tool. In contrast, an execution wrapper that bundles this API with multi-step orchestration, state tracking, and recovery logic (such as inspecting HTTP status codes, falling back to cached archive endpoints upon request timeouts, and parsing DOM trees) autonomously manages procedural control flow, elevating it to a Tool-Execution Skill \citep{song2024easytool, chen2026skillcraft}.
    \item \textbf{Workflows and Frameworks vs. Skills:} The primary cognitive architecture of an agent harness, such as a hardcoded ReAct or Plan-and-Solve control loop, constitutes runtime infrastructure. However, when a composite Directed Acyclic Graph of sub-actions (such as compiling code, running unit test suites, parsing execution tracebacks, and formatting patch diffs) is serialized as an external asset, stored in a repository, and retrieved dynamically to solve a sub-goal, it functions as a modular Composite Skill \citep{taskbench2024, xia2026graspgraphstructuredskillcompositions}.
    \item \textbf{Episodic Memory and Plans vs. Skills:} Episodic memory acts as a passive historical log of past trajectory observations $\mathcal{E}$ without active execution logic, whereas transient plans represent ephemeral, single-use scratchpads generated for the current context. Skills, by contrast, are persistent, generalized, and reusable procedural templates that endure across multi-session lifetimes \citep{packer2023memgpt, sumers2023cognitive}.
\end{itemize}

A systematic comparison of agentic skills against these alternate computational abstractions is summarized in Table~\ref{tab:abstractions}. As shown, skills represent the unique abstraction that unites persistence across resets, modular external execution, runtime composability, and non-parametric modifiability.

\begin{table*}[!htbp]
\centering
\caption{\textbf{Comparison of Computational Abstractions (\S\ref{sec:intro_abstractions}).} Skills are the only abstraction that combines persistence, external execution, runtime composition, and modification without retraining; however, unlike workflows and typed tools, they cannot be fully verified before invocation, which is the source of the security properties analyzed in \S\ref{sec:security_gov}.}
\label{tab:abstractions}
\scriptsize

\setlength{\tabcolsep}{2.5pt}
\renewcommand{\arraystretch}{1.12}
\begin{tabular}{>{\raggedright\arraybackslash}p{1.7cm} >{\raggedright\arraybackslash}p{2.1cm} >{\centering\arraybackslash}p{1.1cm} >{\raggedright\arraybackslash}p{1.6cm} >{\centering\arraybackslash}p{1.2cm} >{\centering\arraybackslash}p{1.2cm} >{\centering\arraybackslash}p{1.2cm} >{\centering\arraybackslash}p{1.2cm} >{\raggedright\arraybackslash}p{1.8cm} >{\raggedright\arraybackslash}p{2.3cm}}
\toprule
\textbf{Abstraction} & \textbf{Components Instantiated} & \textbf{Persists across resets} & \textbf{Invoked by} & \textbf{Executes outside} & \textbf{Composable at runtime} & \textbf{Modifiable w/o retrain} & \textbf{Verifiable before use} & \textbf{Cost per invocation} & \textbf{Reference} \\
\midrule
\textbf{System prompt} & $\mathcal{I}$ & \texttimes & always-on & \texttimes & \texttimes & \checkmark & \texttimes & full, every turn & \cite{yao2022react, wei2022chain, sumers2023cognitive} \\
\textbf{Episodic memory} & $\mathcal{E}$ (traces only) & \checkmark & similarity match & \texttimes & \texttimes & \checkmark & \texttimes & retrieval only & \cite{packer2023memgpt, park2023generativeagents, zhou2026memento} \\
\textbf{Workflow / DAG} & $\mathcal{I}, \pi$ ($\mathcal{A}$ fixed at authoring) & \checkmark & scheduler & partial & \texttimes\ (fixed graph) & \texttimes & \checkmark & per node & \cite{hong2023metagpt, wu2023autogen, li2023chatdev} \\
\textbf{Tool / API} & $\mathcal{T}, \mathcal{E}$ & \checkmark & model emits call & \checkmark & \checkmark\ (stateless) & \texttimes\ (server-side) & \checkmark\ (schema) & per call & \cite{schick2023toolformer, patil2023gorilla, song2024easytool} \\
\textbf{Agentic skill} & $\mathcal{A}, \mathcal{I}, \mathcal{C}, \mathcal{T}, \pi, \mathcal{E}$ & \checkmark & conditional on state & \checkmark & \checkmark & \checkmark & partial & metadata, then body & \cite{wang2023voyager, yang2026autoskill, lu2026contractskillrepairablecontractbasedskills, ling2026agent} \\
\bottomrule
\end{tabular}
\end{table*}

\subsection{Origins of Skill-Based Agent Design}
The development of agentic skills builds upon a technical evolution in model augmentation. The first phase involved static prompting paradigms like ReAct, SayCan, and Inner Monologue, demonstrating that models could perform multi-step reasoning \cite{sumers2023cognitive}. The second phase introduced atomic tool-calling frameworks such as Toolformer, Gorilla, MRKL, APIBank, HuggingGPT, and ViperGPT, prompting language models to emit tokens representing single API calls \cite{schick2023toolformer, patil2023gorilla}. The third phase transitioned to programmatic loops and persistent state-tracking, as seen in Voyager, MemGPT, Generative Agents, Reflexion, and Self-Refine \cite{wang2023voyager, packer2023memgpt}. Finally, the modern fourth phase establishes packaged agent skills as a highly capable abstraction for platforms like SWE-Agent, MetaGPT, AutoGPT, WebArena, and Mind2Web, enabling the secure deployment of pre-compiled directories \cite{taskbench2024}.

\subsection{Summary of Contributions}
This paper provides a unified systems-level foundation for the emerging landscape of agentic skills, formalizing a rapidly growing paradigm into an end-to-end, lifecycle-oriented architecture. In summary, our core contributions are:
\begin{enumerate}
    \item \textbf{Formal Foundations and Mathematical Abstraction:} We formalize agentic skills under a unified six-tuple abstraction $s = (\mathcal{A}, \mathcal{I}, \mathcal{C}, \mathcal{T}, \pi, \mathcal{E})$, establishing a clear conceptual boundary that separates externalized procedural knowledge from static prompting, episodic memories, and stateless API calls (Table~\ref{tab:abstractions}).
    \item \textbf{Procedural Skill Reference Architecture:} We synthesize the procedural lifecycle across nine core phases (Discovery, Representation, Memory Storage, Dynamic Routing, Orchestration, Execution/Repair, Adaptation, Evaluation, and Security Governance), supported by ten comprehensive architectural comparison tables.
    \item \textbf{Taxonomy of Capabilities and Domains:} We classify skill implementations across functional categories (Instructional SOPs, Tool Calling, Latent Reasoning, Meta-Skills) and four operational environments: Software Engineering, GUI/OS Navigation, Embodied Robotics, and Scientific Discovery (Table~\ref{tab:capabilities_domains}).
    \item \textbf{Ecosystems, Marketplaces, and Threat Modeling:} We analyze real-world marketplace dynamics, public skill registries, empirical execution suites, and an adversarial threat taxonomy with corresponding runtime defenses.
\end{enumerate}

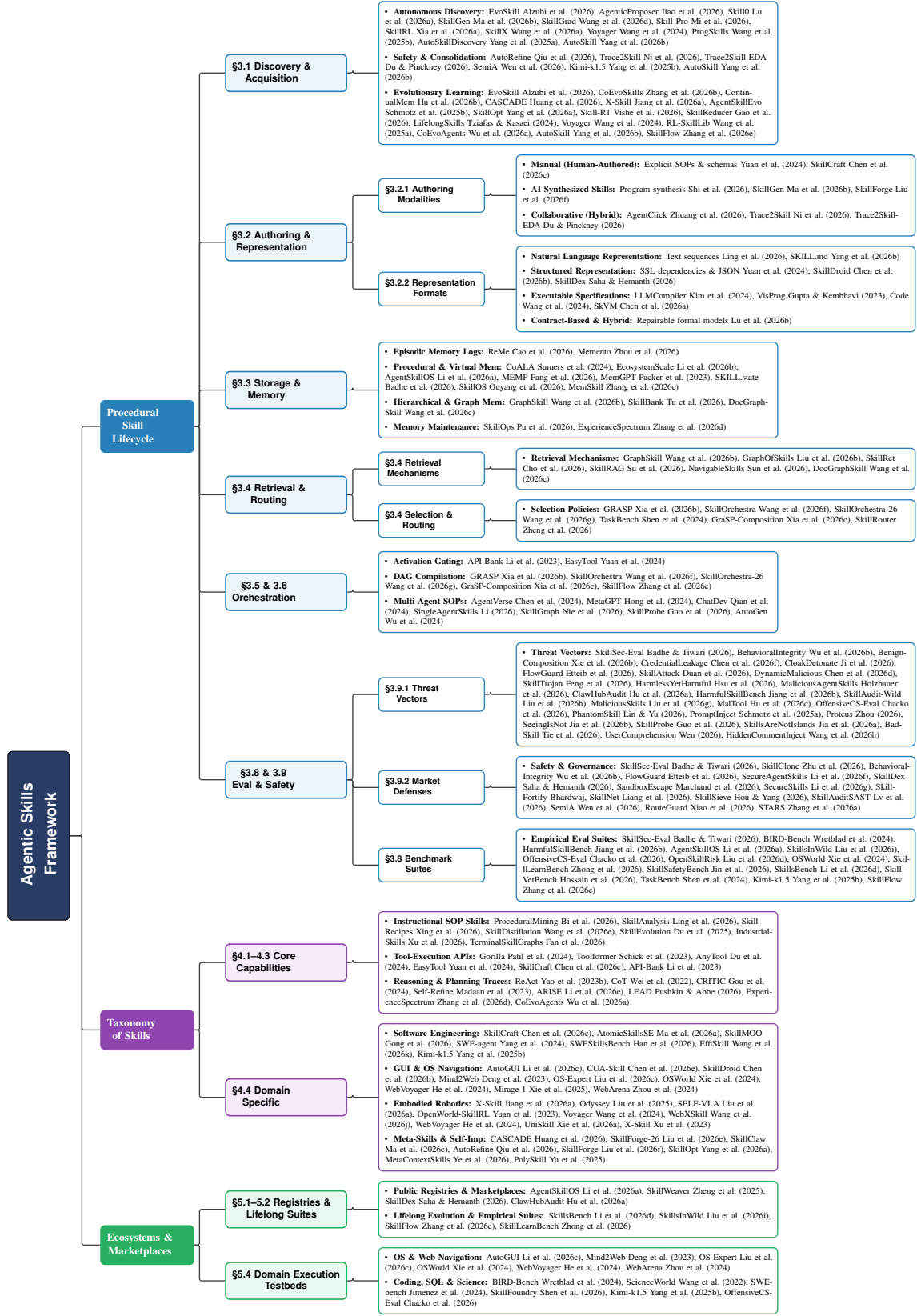
\begin{figure*}[tp]
  \centering
  \begin{adjustbox}{max width=\linewidth, max height=0.95\textheight, keepaspectratio}
    \begin{forest}
      for tree={
        grow=0,                 
        reversed,               
        forked edges,           
        draw,
        rounded corners=4pt,
        node options={inner sep=6pt, align=left},
        l sep=10mm,             
        s sep=3mm,              
        edge={draw=darkgray, line width=1pt, rounded corners=2pt},
        font=\sffamily,
      },
      root/.style={
        fill=colRoot, text=white,
        inner sep=12pt, align=center, anchor=west
      },
      layer/.style 2 args={
        fill=#1, draw=#1, text=white, font=\bfseries\normalsize,
        text width=2.6cm, align=center, anchor=west,
        edge path={
          \noexpand\path[\forestoption{edge}]
            (!u.parent anchor) -- ++(4mm,0) |- (.child anchor)\forestoption{edge label};
        }
      },
      sublayer/.style 2 args={
        fill=#2, draw=#1, line width=1.2pt, text=black,
        font=\bfseries\small\sffamily, text width=3.5cm,
        align=center, anchor=west
      },
      subsublayer/.style 2 args={
        fill=white, draw=#1, line width=1pt, text=black,
        font=\bfseries\footnotesize\sffamily, text width=3.0cm,
        align=center, anchor=west
      },
      leaf/.style 2 args={
        fill=white, draw=#1, line width=1pt, text=black,
        font=\footnotesize\rmfamily, text width=12.5cm, anchor=west
      }
      [{\rotatebox{90}{\parbox{4.5cm}{\centering\Large\bfseries\sffamily Agentic Skills\\[4pt]Framework}}}, root
        [{Procedural\\Skill\\Lifecycle}, layer={colLife}{bgLife}
          [{\S\ref{sec:discovery} Discovery \&\\Acquisition}, sublayer={colLife}{bgLife}
            [{%
              \textbullet~ \textbf{Autonomous Discovery:} EvoSkill \cite{alzubi2026evoskillautomatedskilldiscovery}, AgenticProposer \cite{jiao2026agenticproposingenhancinglarge}, Skill0 \cite{lu2026skill0}, SkillGen \cite{skillgen2026}, SkillGrad \cite{skillgrad2026}, Skill-Pro \cite{skillpro2026}, SkillRL \cite{skillrl2026}, SkillX \cite{skillx2026auto}, Voyager \cite{wang2023voyager}, ProgSkills \cite{wang2025inducing}, AutoSkillDiscovery \cite{yang2025automatedskilldiscoverylanguage}, AutoSkill \cite{yang2026autoskill}\itembreak
              \textbullet~ \textbf{Safety \& Consolidation:} AutoRefine \cite{qiu2026autorefinetrajectoriesreusableexpertise}, Trace2Skill \cite{trace2skill2026}, Trace2Skill-EDA \cite{trace2skilleda2026}, SemiA \cite{wen2026semia}, Kimi-k1.5 \cite{yang2025kimi}, AutoSkill \cite{yang2026autoskill}\itembreak
              \textbullet~ \textbf{Evolutionary Learning:} EvoSkill \cite{alzubi2026evoskillautomatedskilldiscovery}, CoEvoSkills \cite{coevoskills2026}, ContinualMem \cite{continuallearningmemory2026}, CASCADE \cite{huang2026cascadecumulativeagenticskill}, X-Skill \cite{jiang2026xskill}, AgentSkillEvo \cite{schmotz2025agent}, SkillOpt \cite{skillopt2026}, Skill-R1 \cite{skillr12026}, SkillReducer \cite{skillreducer2026}, LifelongSkills \cite{tziafas2024lifelong}, Voyager \cite{wang2023voyager}, RL-SkillLib \cite{wang2025reinforcement}, CoEvoAgents \cite{wu2026coevolvingllmdecisionskill}, AutoSkill \cite{yang2026autoskill}, SkillFlow \cite{zhang2026skillflow}%
            }, leaf={colLife}{bgLife}]
          ]
          [{\S\ref{sec:authoring} Authoring \&\\Representation}, sublayer={colLife}{bgLife}
            [{\S\ref{sec:authoring_mod} Authoring\\Modalities}, subsublayer={colLife}{bgLife}
              [{%
                \textbullet~ \textbf{Manual (Human-Authored):} Explicit SOPs \& schemas \cite{song2024easytool}, SkillCraft \cite{chen2026skillcraft}\itembreak
                \textbullet~ \textbf{AI-Synthesized Skills:} Program synthesis \cite{psn2025programming}, SkillGen \cite{skillgen2026}, SkillForge \cite{skillforge2026}\itembreak
                \textbullet~ \textbf{Collaborative (Hybrid):} AgentClick \cite{zhuang2026agentclickskillbasedhumanintheloopreview}, Trace2Skill \cite{trace2skill2026}, Trace2Skill-EDA \cite{trace2skilleda2026}%
              }, leaf={colLife}{bgLife}]
            ]
            [{\S\ref{sec:authoring_rep} Representation\\Formats}, subsublayer={colLife}{bgLife}
              [{%
                \textbullet~ \textbf{Natural Language Representation:} Text sequences \cite{ling2026agent}, SKILL.md \cite{yang2026autoskill}\itembreak
                \textbullet~ \textbf{Structured Representation:} SSL dependencies \& JSON \cite{song2024easytool}, SkillDroid \cite{chen2026skilldroid}, SkillDex \cite{saha2026skilldex}\itembreak
                \textbullet~ \textbf{Executable Specifications:} LLMCompiler \cite{llmcompiler2023}, VisProg \cite{visprog2023}, Code \cite{wang2023voyager}, SkVM \cite{skillrt2026}\itembreak
                \textbullet~ \textbf{Contract-Based \& Hybrid:} Repairable formal models \cite{lu2026contractskillrepairablecontractbasedskills}%
              }, leaf={colLife}{bgLife}]
            ]
          ]
          [{\S\ref{sec:storage} Storage \&\\Memory}, sublayer={colLife}{bgLife}
            [{%
              \textbullet~ \textbf{Episodic Memory Logs:} ReMe \cite{reme2025}, Memento \cite{zhou2026memento}\itembreak
              \textbullet~ \textbf{Procedural \& Virtual Mem:} CoALA \cite{sumers2023cognitive}, EcosystemScale \cite{ecosystemscale2026}, AgentSkillOS \cite{li2025agentskillos}, MEMP \cite{memp2025}, MemGPT \cite{packer2023memgpt}, SKILL.state \cite{badhe2026skillstate}, SkillOS \cite{skillos2026}, MemSkill \cite{zhang2026memskill}\itembreak
              \textbullet~ \textbf{Hierarchical \& Graph Mem:} GraphSkill \cite{fali2026graphskill}, SkillBank \cite{tu2026dynamicdualgranularityskillbank}, DocGraphSkill \cite{wang2026graphskilldocumentationguidedhierarchicalretrievalaugmented}\itembreak
              \textbullet~ \textbf{Memory Maintenance:} SkillOps \cite{skillops2026}, ExperienceSpectrum \cite{zhang2026experiencecompressionspectrumunifying}%
            }, leaf={colLife}{bgLife}]
          ]
          [{\S\ref{sec:retrieval} Retrieval \&\\Routing}, sublayer={colLife}{bgLife}
            [{\S\ref{sec:retrieval} Retrieval\\Mechanisms}, subsublayer={colLife}{bgLife}
              [{%
                \textbullet~ \textbf{Retrieval Mechanisms:} GraphSkill \cite{fali2026graphskill}, GraphOfSkills \cite{liu2026graphskillsdependencyawarestructural}, SkillRet \cite{skillret2026}, SkillRAG \cite{skillretrievalaugmentationforagenticai2026}, NavigableSkills \cite{sun2026dontretrievenavigatedistilling}, DocGraphSkill \cite{wang2026graphskilldocumentationguidedhierarchicalretrievalaugmented}%
              }, leaf={colLife}{bgLife}]
            ]
            [{\S\ref{sec:retrieval} Selection \&\\Routing}, subsublayer={colLife}{bgLife}
              [{%
                \textbullet~ \textbf{Selection Policies:} GRASP \cite{grasp2026}, SkillOrchestra \cite{skillorchestra2025}, SkillOrchestra-26 \cite{skillorchestra2026}, TaskBench \cite{taskbench2024}, GraSP-Composition \cite{xia2026graspgraphstructuredskillcompositions}, SkillRouter \cite{zheng2026skillrouter}%
              }, leaf={colLife}{bgLife}]
            ]
          ]
          [{\S\ref{sec:composition} \& \ref{sec:execution}\\Orchestration}, sublayer={colLife}{bgLife}
            [{%
              \textbullet~ \textbf{Activation Gating:} API-Bank \cite{apibank2023}, EasyTool \cite{song2024easytool}\itembreak
              \textbullet~ \textbf{DAG Compilation:} GRASP \cite{grasp2026}, SkillOrchestra \cite{skillorchestra2025}, SkillOrchestra-26 \cite{skillorchestra2026}, GraSP-Composition \cite{xia2026graspgraphstructuredskillcompositions}, SkillFlow \cite{zhang2026skillflow}\itembreak
              \textbullet~ \textbf{Multi-Agent SOPs:} AgentVerse \cite{agentverse2023}, MetaGPT \cite{hong2023metagpt}, ChatDev \cite{li2023chatdev}, SingleAgentSkills \cite{li2026single}, SkillGraph \cite{skillgraph2026}, SkillProbe \cite{skillprobe2026}, AutoGen \cite{wu2023autogen}%
            }, leaf={colLife}{bgLife}]
          ]
          [{\S\ref{sec:eval_suites} \& \ref{sec:security_gov}\\Eval \& Safety}, sublayer={colLife}{bgLife}
            [{\S\ref{sec:threat_vectors} Threat\\Vectors}, subsublayer={colLife}{bgLife}
              [{%
                \textbullet~ \textbf{Threat Vectors:} SkillSec-Eval \cite{badhe2025agent}, BehavioralIntegrity \cite{behavioral2026integrity}, BenignComposition \cite{benigncomposition2026}, CredentialLeakage \cite{chen2026credentialleakagellmagent}, CloakDetonate \cite{cloakdetonate2026}, FlowGuard \cite{detectingmalicious2026}, SkillAttack \cite{duan2026skillattack}, DynamicMalicious \cite{dynamicmalicious2026}, SkillTrojan \cite{feng2026skilltrojan}, HarmlessYetHarmful \cite{harmlessyetharmful2026}, MaliciousAgentSkills \cite{holzbauer2026malicious}, ClawHubAudit \cite{hu2026red}, HarmfulSkillBench \cite{jiang2026harmfulskillbenchharmfulskillsweaponize}, SkillAudit-Wild \cite{liu2026agentskillswildempirical}, MaliciousSkills \cite{liu2026malicious}, MalTool \cite{maltool2026}, OffensiveCS-Eval \cite{negativeresult2026}, PhantomSkill \cite{phantomskill2026}, PromptInject \cite{promptinject2025skills}, Proteus \cite{proteus2026selfevolving}, SeeingIsNot \cite{seeingisnot2026}, SkillProbe \cite{skillprobe2026}, SkillsAreNotIslands \cite{skillsarenot2026}, BadSkill \cite{tie2026badskillbackdoorattacksagent}, UserComprehension \cite{usercomprehension2026}, HiddenCommentInject \cite{wang2026skills}%
              }, leaf={colLife}{bgLife}]
            ]
            [{\S\ref{sec:market_defenses} Market\\Defenses}, subsublayer={colLife}{bgLife}
              [{%
                \textbullet~ \textbf{Safety \& Governance:} SkillSec-Eval \cite{badhe2025agent}, SkillClone \cite{zhu2026skillclone}, BehavioralIntegrity \cite{behavioral2026integrity}, FlowGuard \cite{detectingmalicious2026}, SecureAgentSkills \cite{li2026towards}, SkillDex \cite{saha2026skilldex}, SandboxEscape \cite{sandboxescape2026}, SecureSkills \cite{secureskills2026}, SkillFortify \cite{skillfortify2026}, SkillNet \cite{skillnet2026}, SkillSieve \cite{skillsieve2026}, SkillAuditSAST \cite{structuredsecurityauditingandrobustnessenhancementforuntrustedagentskills2026}, SemiA \cite{wen2026semia}, RouteGuard \cite{xiao2026routeguard}, STARS \cite{zhang2026stars}%
              }, leaf={colLife}{bgLife}]
            ]
            [{\S\ref{sec:eval_suites} Benchmark\\Suites}, subsublayer={colLife}{bgLife}
              [{%
                \textbullet~ \textbf{Empirical Eval Suites:} SkillSec-Eval \cite{badhe2025agent}, BIRD-Bench \cite{bird2023}, HarmfulSkillBench \cite{jiang2026harmfulskillbenchharmfulskillsweaponize}, AgentSkillOS \cite{li2025agentskillos}, SkillsInWild \cite{liu2026agenticskillsworkwild}, OffensiveCS-Eval \cite{negativeresult2026}, OpenSkillRisk \cite{openskillrisk2026}, OSWorld \cite{osworld2024}, SkillLearnBench \cite{skilllearnbench2026}, SkillSafetyBench \cite{skillsafetybench2026}, SkillsBench \cite{skillsbench2026}, SkillVetBench \cite{skillvetbench2026}, TaskBench \cite{taskbench2024}, Kimi-k1.5 \cite{yang2025kimi}, SkillFlow \cite{zhang2026skillflow}%
              }, leaf={colLife}{bgLife}]
            ]
          ]
        ]
        [{Taxonomy\\of Skills}, layer={colTax}{bgTax}
          [{\S\ref{sec:instructional}--\ref{sec:reasoning} Core\\Capabilities}, sublayer={colTax}{bgTax}
            [{%
              \textbullet~ \textbf{Instructional SOP Skills:} ProceduralMining \cite{procedural2024}, SkillAnalysis \cite{ling2026agent}, SkillRecipes \cite{xing2026recipes}, SkillDistillation \cite{wang2026skill}, SkillEvolution \cite{du2025rethinking}, IndustrialSkills \cite{xu2026agentskillframeworkperspectives}, TerminalSkillGraphs \cite{fan2026toward}\itembreak
              \textbullet~ \textbf{Tool-Execution APIs:} Gorilla \cite{patil2023gorilla}, Toolformer \cite{schick2023toolformer}, AnyTool \cite{anytool2024}, EasyTool \cite{song2024easytool}, SkillCraft \cite{chen2026skillcraft}, API-Bank \cite{apibank2023}\itembreak
              \textbullet~ \textbf{Reasoning \& Planning Traces:} ReAct \cite{yao2022react}, CoT \cite{wei2022chain}, CRITIC \cite{critic2023}, Self-Refine \cite{madaan2023selfrefine}, ARISE \cite{li2026ariseagentreasoningintrinsic}, LEAD \cite{norecoverybottleneck2026}, ExperienceSpectrum \cite{zhang2026experiencecompressionspectrumunifying}, CoEvoAgents \cite{wu2026coevolvingllmdecisionskill}%
            }, leaf={colTax}{bgTax}]
          ]
          [{\S\ref{sec:domain_specific} Domain\\Specific}, sublayer={colTax}{bgTax}
            [{%
              \textbullet~ \textbf{Software Engineering:} SkillCraft \cite{chen2026skillcraft}, AtomicSkillsSE \cite{ma2026scaling}, SkillMOO \cite{skillmoo2026}, SWE-agent \cite{sweagent2024}, SWESkillsBench \cite{sweskillsbench2026}, EffiSkill \cite{wang2026effiskillagentskillbased}, Kimi-k1.5 \cite{yang2025kimi}\itembreak
              \textbullet~ \textbf{GUI \& OS Navigation:} AutoGUI \cite{autogui2023}, CUA-Skill \cite{chen2026cuaskilldevelopskillscomputer}, SkillDroid \cite{chen2026skilldroid}, Mind2Web \cite{deng2023mind2web}, OS-Expert \cite{liu2026osexpert}, OSWorld \cite{osworld2024}, WebVoyager \cite{webvoyager2024}, Mirage-1 \cite{xie2025mirage}, WebArena \cite{zhou2023webarena}\itembreak
              \textbullet~ \textbf{Embodied Robotics:} X-Skill \cite{jiang2026xskill}, Odyssey \cite{liu2025odysseyempoweringminecraftagents}, SELF-VLA \cite{liu2026selfvlaskillenhancedagentic}, OpenWorld-SkillRL \cite{skillreinforcementlearningandplanningforopenworldlonghorizontasks2026}, Voyager \cite{wang2023voyager}, WebXSkill \cite{wang2026webxskill}, WebVoyager \cite{webvoyager2024}, UniSkill \cite{xie2026uniskillbuildingselfevolvingskill}, X-Skill \cite{xu2023xskillcrossembodimentskill}\itembreak
              \textbullet~ \textbf{Meta-Skills \& Self-Imp:} CASCADE \cite{huang2026cascadecumulativeagenticskill}, SkillForge-26 \cite{liu2026skillforge}, SkillClaw \cite{ma2026skillclaw}, AutoRefine \cite{qiu2026autorefinetrajectoriesreusableexpertise}, SkillForge \cite{skillforge2026}, SkillOpt \cite{skillopt2026}, MetaContextSkills \cite{ye2026meta}, PolySkill \cite{yu2025polyskill}%
            }, leaf={colTax}{bgTax}]
          ]
        ]
        [{Ecosystems \&\\Marketplaces}, layer={colEco}{bgEco}
          [{\S\ref{sec:public_registries}--\ref{sec:lifelong_ecosystems} Registries \&\\Lifelong Suites}, sublayer={colEco}{bgEco}
            [{%
              \textbullet~ \textbf{Public Registries \& Marketplaces:} AgentSkillOS \cite{li2025agentskillos}, SkillWeaver \cite{zheng2025skillweaver}, SkillDex \cite{saha2026skilldex}, ClawHubAudit \cite{hu2026red}\itembreak
              \textbullet~ \textbf{Lifelong Evolution \& Empirical Suites:} SkillsBench \cite{skillsbench2026}, SkillsInWild \cite{liu2026agenticskillsworkwild}, SkillFlow \cite{zhang2026skillflow}, SkillLearnBench \cite{skilllearnbench2026}%
            }, leaf={colEco}{bgEco}]
          ]
          [{\S\ref{sec:domain_execution_suites} Domain Execution\\Testbeds}, sublayer={colEco}{bgEco}
            [{%
              \textbullet~ \textbf{OS \& Web Navigation:} AutoGUI \cite{autogui2023}, Mind2Web \cite{deng2023mind2web}, OS-Expert \cite{liu2026osexpert}, OSWorld \cite{osworld2024}, WebVoyager \cite{webvoyager2024}, WebArena \cite{zhou2023webarena}\itembreak
              \textbullet~ \textbf{Coding, SQL \& Science:} BIRD-Bench \cite{bird2023}, ScienceWorld \cite{scienceworld2021}, SWE-bench \cite{swebench2024}, SkillFoundry \cite{skillfoundry2026}, Kimi-k1.5 \cite{yang2025kimi}, OffensiveCS-Eval \cite{negativeresult2026}%
            }, leaf={colEco}{bgEco}]
          ]
        ]
      ] 
    \end{forest}
  \end{adjustbox}
  \caption{Architectural Blueprint and Systems Landscape of the Agentic Skills Framework, categorizing key system methodologies, lifecycle stages, and domain environments.}
  \label{fig:skills_framework_deep}
\end{figure*}

\section{Foundations of Skill-Based Agents}

\subsection{Cognitive Architectures and Procedural Memory}
In cognitive psychology, memory is broadly divided into declarative memory, which handles semantic facts, and procedural memory, which governs the procedural mechanics of task execution \cite{sumers2023cognitive}. Computational cognitive architectures model human problem-solving as a tight, iterative loop where declarative knowledge is compiled into procedural rules over time \cite{sumers2023cognitive}. Within the context of language agents, agentic skills serve as the non-parametric procedural memory of the autonomous system. While parametric model weights represent a static semantic memory, and vector databases represent a transient episodic memory, skills provide the modular execution substrate required for repeatable tasks \cite{packer2023memgpt}. When faced with a familiar problem, the agent retrieves the corresponding procedural skill and executes it directly, bypassing slow reasoning steps \cite{sumers2023cognitive}.

\subsection{Skills as Externalized Procedural Knowledge}
The core architectural innovation lies in the externalization of this procedural memory. Attempting to feed extensive procedural instructions directly into the context window inflates inference latency and degrades accuracy \cite{negativeresult2026}. Externalizing procedural knowledge into packaged files resolves these limitations. Furthermore, externalization enables the system-level principle of progressive disclosure, where the agent reads only the lightweight metadata first, loading full implementation code only during active execution \cite{song2024easytool}. Because these skills are stored as plain-text assets, they can be updated without requiring expensive model retraining loops \cite{zhuang2026agentclickskillbasedhumanintheloopreview}.

We formalize the core primitives of this framework under a unified mathematical abstraction:

\textbf{Definition 1 (Skill):} A skill $s$ is a reusable procedural abstraction consisting of activation conditions ($\mathcal{A}$), procedural instructions ($\mathcal{I}$), execution constraints ($\mathcal{C}$), tool interfaces ($\mathcal{T}$), execution policy ($\pi$), and intended effects ($\mathcal{E}$) \cite{song2024easytool}:
\begin{equation}
s = (\mathcal{A}, \mathcal{I}, \mathcal{C}, \mathcal{T}, \pi, \mathcal{E})
\end{equation}

\textbf{Definition 2 (Agent State):} The agent state at time $t$ is denoted as $x_t \in \mathcal{X}$, capturing conversation history, internal memory, environment observations, tool outputs, active goals $g \in \mathcal{G}$, and retrieved documents \cite{packer2023memgpt}.

\textbf{Definition 3 (Skill Activation):} A skill activates conditionally based on the agent state $x_t$ and the active task goal $g$:
\begin{equation}
\mathcal{A}(x_t, g) \rightarrow \{0, 1\}
\end{equation}
where $\mathcal{A}(x_t, g) = 1$ indicates that the skill should trigger, formalizing dynamic routing, contextual applicability, and planner-guided invocation \cite{zheng2026skillrouter}. Crucially, to prevent unauthorized execution and privilege escalation, activation is governed by a capability-based permission constraint:
\begin{equation}
\mathcal{A}(x_t, g) = 1 \;\Rightarrow\; \mathcal{T}_s \subseteq \mathcal{T}_{\text{permitted}}(x_t)
\end{equation}
A skill may fire only if its declared tool set $\mathcal{T}_s$ is a strict subset of the tools permitted by the active execution context $\mathcal{T}_{\text{permitted}}(x_t)$ \cite{secureskills2026}. This containment formalizes the capability-based permission model required for secure agent execution, defining privilege escalation as any violation of this subset condition ($\mathcal{T}_s \not\subseteq \mathcal{T}_{\text{permitted}}(x_t)$), and establishing a formal boundary between runtime sandboxing and pre-deployment static auditing.

\textbf{Definition 4 (Skill Execution):} When activated, a skill induces a conditional state transition:
\begin{equation}
x_{t+1} = s(x_t) = \pi(x_t, \mathcal{I}, \mathcal{T}) \quad \text{if} \quad \mathcal{A}(x_t, g) = 1
\end{equation}
This formalizes procedural execution, deterministic tool interaction, and reasoning under conditional invocation \cite{zheng2026skillrouter}.

\section{Procedural Skill Lifecycle in LLM Agents}\label{sec:lifecycle}
The procedural skill lifecycle represents the core operational architecture of modern agent platforms, governing how skills transition from initial creation to runtime deployment, adaptation, and eventual decommissioning. Managing these stages dynamically is critical to maintaining a reliable, safe, and efficient agent capability ecosystem \cite{yang2026autoskill}. This section details each stage of the lifecycle, unifying them under a consistent mathematical framework. While this section details the operational lifecycle of skills, a comprehensive functional classification across domains is provided in Section~\ref{sec:taxonomy} (Table~\ref{tab:capabilities_domains}), and an analysis of skill ecosystems and marketplaces is provided in Section~\ref{sec:ecosystems} (Table~\ref{tab:ecosystems_benchmarks}).

\subsection{Skill Discovery and Acquisition}\label{sec:discovery}
When deployed in open-ended or unfamiliar operational environments, autonomous language agents cannot rely exclusively on fixed, pre-compiled skill libraries. To achieve genuine autonomy, agents require mechanisms to autonomously discover, formalize, and consolidate new procedural capabilities from ongoing interactive experience \citep{yang2026autoskill, coevoskills2026}. As summarized in Table~\ref{tab:discovery_authoring}, contemporary frameworks approach discovery from fundamentally different assumptions regarding trajectory sources, failure signal utilization, and pre-admission verification rigor.

\begin{table*}[!htbp]
\centering
\caption{\textbf{Discovery, Acquisition \& Authoring (\S\ref{sec:discovery} \& \S\ref{sec:authoring_mod}).} Comparison of procedural skill creation and evolution frameworks ($n=19$ systems). Sorted primarily by \textbf{Authoring modality} (\textit{Autonomous RL} $\rightarrow$ \textit{AI-synthesized} $\rightarrow$ \textit{Hybrid} $\rightarrow$ \textit{Manual}), then secondarily by \textbf{Updates over time}. \textbf{Columns:} \textbf{System} (procedural creation/evolution framework); \textbf{Core Contribution} (primary algorithmic or structural innovation); \textbf{Trajectory Source} (input origin of procedural knowledge: dialogue, task rollout, embodied, or manual); \textbf{Authoring Modality} (abstraction generation mechanism: autonomous RL, AI-synthesized, hybrid HITL, or manual); \textbf{Failure Signal Usage} (how failure traces inform synthesis: repair signal, contrastive RL pairs, rejection filter, or not used); \textbf{Verified Before Commit} (pre-admission quality gating: formal, unit test, LLM judge, HITL review, or none); \textbf{Updates Over Time} (continual adaptation target: instructions, policy, library membership, or none).}
\label{tab:discovery_authoring}
\scriptsize
\setlength{\tabcolsep}{2.5pt}
\renewcommand{\arraystretch}{1.12}
\begin{tabular}{>{\raggedright\arraybackslash}p{2.0cm} >{\raggedright\arraybackslash}p{3.8cm} >{\raggedright\arraybackslash}p{2.0cm} >{\raggedright\arraybackslash}p{1.8cm} >{\raggedright\arraybackslash}p{2.0cm} >{\raggedright\arraybackslash}p{2.2cm} >{\raggedright\arraybackslash}p{2.4cm}}
\toprule
\textbf{System} & \textbf{Core Contribution} & \textbf{Trajectory Source} & \textbf{Authoring Modality} & \textbf{Failure Signal Usage} & \textbf{Verified Before Commit} & \textbf{Updates Over Time} \\
\midrule
\multicolumn{7}{l}{\textit{\textbf{Autonomous Reinforcement Learning (RL-Induced Procedural Abstractions)}}} \\
\midrule
\textbf{SkillRL} \cite{skillrl2026} & RL-induced procedural policy & Task rollout & Autonomous RL & Contrastive pairs & Unit test & Policy, membership \\
\textbf{Skill-Pro} \cite{skillpro2026} & Non-parametric PPO over Skill-MDP & Task rollout & Autonomous RL & Contrastive pairs & PPO Gate & Membership, instructions \\
\textbf{Skill-R1} \cite{skillr12026} & RL-based skill evolution & Task rollout & Autonomous RL & Contrastive pairs & Unit test & Membership \\
\textbf{Skill0} \cite{lu2026skill0} & In-context RL internalization & Task rollout & Autonomous RL & Contrastive pairs & None & Policy (internalized) \\
\midrule
\multicolumn{7}{l}{\textit{\textbf{AI-Synthesized Skills (LLM-Generated SOPs, Trajectory Consolidation \& Textual Evolution)}}} \\
\midrule
SkillOpt \cite{skillopt2026} & Controllable text-space optimizer & Task rollout & AI-synthesized & Rejection filter & Unit test, LLM judge & Instructions \\
\textbf{Voyager} \cite{wang2023voyager} & Continual embodied skill library & Embodied & AI-synthesized & Repair signal & Unit test & Membership \\
\textbf{AutoSkill} \cite{yang2026autoskill} & Experience-driven lifelong dialogue & Dialogue, rollout & AI-synthesized & Not used & LLM judge & Membership \\
EvoSkill \cite{alzubi2026evoskillautomatedskilldiscovery} & Automated evolutionary discovery & Task rollout & AI-synthesized & Repair signal & LLM judge & Instructions, membership \\
\textbf{ProgSkills} \cite{wang2025inducing} & Programmatic skill induction & Task rollout & AI-synthesized & Not used & Unit test & Membership \\
\textbf{SkillGen} \cite{skillgen2026} & Inference-time verified synthesis & Task rollout & AI-synthesized & Contrastive pairs & Unit test, HITL & Membership \\
\textbf{SkillX} \cite{skillx2026auto} & Automated skill construction & Task rollout & AI-synthesized & Not used & Unit test & Membership \\
\textbf{Trace2Skill} \cite{trace2skill2026} & Distills trajectory failure lessons & Task rollout & AI-synthesized & Repair signal & Unit test & Membership \\
\textbf{Trace2Skill-EDA} \cite{trace2skilleda2026} & Trajectory exploratory analysis & Task rollout & AI-synthesized & Repair signal & Unit test & Membership \\
\textbf{AutoRefine} \cite{qiu2026autorefinetrajectoriesreusableexpertise} & Refining trajectory expertise & Task rollout & AI-synthesized & Repair signal & LLM judge & Membership \\
\textbf{CASCADE} \cite{huang2026cascadecumulativeagenticskill} & Cumulative skill development & Task rollout & AI-synthesized & Repair signal & Unit test & Membership \\
\textbf{CoEvoSkills} \cite{coevoskills2026} & Co-evolving decision \& skill banks & Task rollout & AI-synthesized & Repair signal & LLM judge & Membership, instructions \\
PSN \cite{psn2025programming} & Programmatic skill networks & Embodied & AI-synthesized & Repair signal & Unit test & Membership \\
\midrule
\multicolumn{7}{l}{\textit{\textbf{Hybrid \& Manual Authoring (Human-in-the-Loop Review \& Hand-Authored SOP Wrappers)}}} \\
\midrule
\textbf{AgentClick} \cite{zhuang2026agentclickskillbasedhumanintheloopreview} & Skill-based HITL review & Dialogue & Hybrid & n/a & HITL review & Instructions \\
\textbf{EasyTool} \cite{song2024easytool} & Manual SOP schema wrappers & None (manual) & Manual & n/a & None & None \\
\bottomrule
\end{tabular}
\vspace{4pt}
\begin{minipage}{\linewidth}
\scriptsize
\textbf{Notes $\cdot$ Column Value Definitions.} 
\textbf{Trajectory Source:} Origin of procedural experience (\textit{Task rollout}: agent-environment interaction traces; \textit{Dialogue}: multi-turn conversational logs; \textit{Embodied}: physical/3D simulator actions; \textit{None (manual)}: human-authored from scratch). 
\textbf{Authoring Modality:} Creation mechanism (\textit{Autonomous RL}: reinforcement learning optimization; \textit{AI-synthesized}: LLM-generated code/SOPs from trajectories; \textit{Hybrid}: AI synthesis with Human-in-the-Loop review; \textit{Manual}: human engineering). 
\textbf{Failure Signal Usage:} How failed rollouts inform synthesis (\textit{Contrastive pairs}: success/failure trajectories used as RL reward signals; \textit{Repair signal}: error tracebacks fed back to LLM for iterative correction; \textit{Rejection filter}: failed rollouts discarded without learning; \textit{Not used}: only successful trajectories retained). 
\textbf{Verified Before Commit:} Pre-admission quality gating (\textit{Unit test}: executable validation against test suites; \textit{LLM judge}: model-based utility/schema inspection; \textit{HITL review}: human review before commit; \textit{None}: unverified admission). 
\textbf{Updates Over Time:} Target of continual adaptation (\textit{Instructions}: textual SOP/prompt refinement; \textit{Policy}: parametric weight or RL policy updates; \textit{Membership}: adding/pruning skills in the library; \textit{None}: static library).
\end{minipage}
\end{table*}

\subsubsection{Formalizing the Discovery and Acquisition Pipeline}
We formalize skill discovery as the mathematical problem of extracting reusable sub-policies from a distribution of unstructured, long-horizon rollout logs $\mathcal{H} = \{\tau^{(1)}, \tau^{(2)}, \dots, \tau^{(N)}\}$, where each trajectory $\tau = (x_1, a_1, r_1, \dots, x_H)$ records sequences of environment observations, actions, intermediate rewards, and conversational context \citep{coevoskills2026, yang2025automatedskilldiscoverylanguage}. The discovery operator $\mathbb{D}$ maps the trajectory history $\mathcal{H}$ to a candidate set of unverified skill representations $\mathcal{S}_{\text{new}}$:
\begin{equation}
\mathcal{S}_{\text{new}} = \mathbb{D}(\mathcal{H})
\end{equation}
In practice, $\mathbb{D}$ operates by identifying an optimal sub-trajectory $\tau_{i:j} \subseteq \tau$ (with $1 \le i < j \le H$) that satisfies two conditions: (1) \emph{persistent environmental outcome}, inducing a predictable state transformation $\mathcal{E}$, and (2) \emph{entropy minimization}, reducing the branching factor of future policy decisions \citep{skillrl2026, shu2017hierarchicalinterpretableskillacquisition}. Once a sub-trajectory is partitioned, the agent induces an applicability context $\mathcal{C} \subseteq \mathcal{X}$ defining the region of state space where the procedural routine remains numerically or semantically stable, which defines the initial triggering condition $\mathcal{A}_{\text{new}}$:
\begin{equation}
\mathcal{A}_{\text{new}}(x_t, g) = 1 \iff x_t \in \mathcal{C}
\end{equation}

Following discovery, the \emph{Skill Acquisition} operator $\mathbb{A}$ refines and consolidates raw candidates $s \in \mathcal{S}_{\text{new}}$ into production-ready skills $s^* = (\mathcal{A}^*, \mathcal{I}^*, \mathcal{C}^*, \mathcal{T}^*, \pi^*, \mathcal{E}^*)$ committed to persistent procedural memory $\mathcal{M}$:
\begin{equation}
\mathcal{M}_{t+1} = \mathcal{M}_t \cup \{s^*\}
\end{equation}
The acquisition pipeline optimizes the instructions $\mathcal{I}^*$ and executable policy $\pi^*$ to maximize expected operational reward under applicability constraints, subject to a pre-admission verification constraint $V(s^*) = 1$:
\begin{equation}
s^* = \arg\max_s \mathbb{E}_{x \sim \mathcal{C}}[R(s, x)] \quad \text{subject to} \quad V(s^*) = 1
\end{equation}
where $V(s^*) \in \{0, 1\}$ formalizes pre-commit verification (e.g., unit test pass rates, static AST analysis, or formal Datalog invariant proving) to guarantee that newly consolidated skills do not introduce syntactic regressions, interface drift, or privilege escalation vulnerabilities \citep{wen2026semia, agentless2024}.

\subsubsection{Comparative Paradigms: Autonomous RL vs. Inductive AI Synthesis}
The literature diverges sharply on how the discovery operator $\mathbb{D}$ and acquisition operator $\mathbb{A}$ are instantiated:

\paragraph{Autonomous Reinforcement Learning (Option Discovery):}
Frameworks such as SkillRL \citep{skillrl2026}, Skill-Pro \citep{skillpro2026}, Skill-R1 \citep{skillr12026}, and Skill0 \citep{lu2026skill0} ground skill discovery in classical hierarchical reinforcement learning and Markov Decision Process (MDP) option discovery. These architectures discover sub-policies by maximizing intrinsic reward signals (e.g., state reachability, bottleneck state visitation, or contrastive advantage) over task rollouts. For instance, SkillRL \citep{skillrl2026} discovers atomic execution options by training policy sub-networks against environment rewards, distilling rollouts through a teacher model into a two-tier \texttt{SKILLBANK} (universal general skills $S_g$ and task-specific skills $S_k$), before co-evolving the library with policy updates via Group Relative Policy Optimization (GRPO). Similarly, Skill-Pro \citep{skillpro2026} formalizes a Skill-MDP where non-parametric Proximal Policy Optimization extracts natural-language semantic gradients ($g_i = \nabla_{\text{sem}}(\tau_i, \omega)$) from trajectory batches, gating candidate skill admissions through a clipped surrogate verification functional (PPO Gate) and maintaining pool efficiency via online advantage-score pruning.
\emph{Underlying Assumption \& Limitation:} Autonomous RL methods assume environment rewards are dense and execution environments are computationally cheap to simulate. While they achieve high execution fidelity in bounded game environments or robotics simulations \citep{11127927, liu2026selfvlaskillenhancedagentic}, their sample complexity ($10^4$--$10^6$ rollout steps) becomes prohibitively expensive when applied to real-world software engineering repositories or API-constrained web agents where every rollout consumes external model tokens.

\paragraph{Inductive AI Synthesis (Code and SOP Induction):}
To overcome the sample inefficiency of RL, architectures such as Voyager \citep{wang2023voyager}, ProgSkills \citep{wang2025inducing}, SkillGen \citep{skillgen2026}, SkillX \citep{skillx2026auto}, and AutoSkill \citep{yang2026autoskill} leverage the in-context reasoning of frontier LLMs to synthesize skill code and markdown SOPs from as few as $1$--$5$ demonstration traces. In Voyager \citep{wang2023voyager}, an iterative prompting mechanism converts interactive Minecraft environment feedback and execution errors into reusable JavaScript control programs stored in a vector database. In dialogue-centric and workflow domains, AutoSkill \citep{yang2026autoskill} continually parses multi-turn conversation trajectories, distilling conversational insights into structured \texttt{SKILL.md} procedural files without human supervision.
\emph{Underlying Assumption \& Limitation:} Inductive synthesis assumes the underlying foundation model possesses sufficient latent programming capabilities to accurately generalize from noisy traces. In practice, unsupervised synthesis without execution grounding frequently produces brittle skills that overfit to idiosyncrasies of the specific rollout trace, suffering from parameter hallucinations and ambiguous activation boundaries.

\paragraph{Evolutionary Search in Text Space:}
Bridging RL and inductive synthesis, EvoSkill \citep{alzubi2026evoskillautomatedskilldiscovery}, SkillOpt \citep{skillopt2026}, and CASCADE \citep{huang2026cascadecumulativeagenticskill} frame skill discovery as evolutionary optimization directly over textual prompt and code spaces. EvoSkill applies genetic operators (mutation, crossover, and fitness selection) over candidate skill instructions, using task pass rates and execution latencies as multi-objective fitness scores. SkillOpt introduces a controllable text-space optimizer that refines instruction formulations using rejection filtering over task rollouts.

\subsubsection{The Role of Failure Signals: Positive-Only vs. Contrastive Mining}
A major point of divergence across discovery architectures is whether and how execution failure traces are utilized:
\begin{itemize}
    \item \textbf{Positive-Only Induction:} Early architectures, including Voyager \citep{wang2023voyager}, ProgSkills \citep{wang2025inducing}, and AutoSkill \citep{yang2026autoskill}, discard failed trajectories entirely, inducing skills strictly from successful rollout trajectories. While computationally straightforward, positive-only induction leaves skills blind to operational failure modes, producing procedures that fail catastrophically when encountering runtime exceptions.
    \item \textbf{Contrastive Mining and Error Tracebacks:} Recent systems, such as Trace2Skill \citep{trace2skill2026}, Trace2Skill-EDA \citep{trace2skilleda2026}, AutoRefine \citep{qiu2026autorefinetrajectoriesreusableexpertise}, CoEvoSkills \citep{coevoskills2026}, and SkillGen \citep{skillgen2026}, demonstrate that analyzing paired success-failure traces significantly improves skill robustness. Trace2Skill \citep{trace2skill2026} deploys asymmetric analyst sub-agents: a success analyst $A^+$ extracts positive execution patterns while an interactive error analyst $A^-$ inspects execution tracebacks, file diffs, and ground-truth mismatches to generate targeted patch proposals. These patches are merged through hierarchical many-to-one consolidation into standard operating procedures (e.g., formula recalculation, write-back verification), drastically improving wall-clock consolidation efficiency over sequential prompt editing. In hardware engineering, Trace2Skill-EDA \citep{trace2skilleda2026} proves that coupling dense verifier feedback with task-wise error distillation substantially boosts complex EDA script execution success. SkillGen \citep{skillgen2026} clusters failure and success summaries via $k$-means embeddings to synthesize contrastive pre-condition boundaries $\mathcal{C}$, preventing the skill from triggering in out-of-distribution states.
\end{itemize}

\subsubsection{Pre-Admission Quality Gating and Verification}
Before a discovered skill candidate is committed to persistent memory $\mathcal{M}$, frameworks enforce varying degrees of quality gating $V(s)$:
\begin{itemize}
    \item \textbf{Unit Testing and Sandbox Execution:} Systems like Voyager \citep{wang2023voyager}, SkillRL \citep{skillrl2026}, ProgSkills \citep{wang2025inducing}, and PSN \citep{psn2025programming} demand that synthesized skills pass deterministically constructed unit tests inside isolated sandbox containers prior to library admission.
    \item \textbf{LLM-as-a-Judge Vetting:} Frameworks operating on natural language SOPs (AutoRefine \citep{qiu2026autorefinetrajectoriesreusableexpertise}, CASCADE \citep{huang2026cascadecumulativeagenticskill}, EvoSkill \citep{alzubi2026evoskillautomatedskilldiscovery}) deploy secondary LLM judges to inspect candidate skill descriptions, evaluating clarity, modularity, and redundancy against the existing skill bank.
    \item \textbf{Formal Contract Proving:} Verification frameworks like SemiA \citep{wen2026semia} translate skill instructions and execution bodies into formal Datalog rules, checking temporal safety invariants before permitting deployment to production registries.
    \item \textbf{Human-in-the-Loop Review:} Collaborative platforms like AgentClick \citep{zhuang2026agentclickskillbasedhumanintheloopreview} inject human verification checkpoints where human engineers approve or modify synthesized procedures before deployment into production workflows.
\end{itemize}

\subsubsection{Critical Methodological Weaknesses and Open Bottlenecks}
Our synthesis identifies three fundamental bottlenecks currently limiting autonomous discovery:
\begin{enumerate}
    \item \textbf{The Exploration-Exploitation Gap in Sparse Environments:} In complex software engineering repositories (e.g., SWE-bench \citep{swebench2024}) or open-world operating systems (OSWorld \citep{osworld2024}), unguided exploration rarely encounters successful end-to-end trajectories by chance. Without human demonstration traces or heavily shaped curricula, autonomous discovery algorithms fail to bridge the initial zero-to-one capability gap.
    \item \textbf{Skill Proliferation and Memory Dilution:} Unchecked autonomous discovery rapidly floods the skill repository with low-utility, overspecialized, or near-duplicate skill files ($n > 10^3$). As demonstrated by \citet{memp2025} and \citet{liu2026agenticskillsworkwild}, large uncurated skill pools saturate retrieval indices, increasing semantic confusability and severely degrading downstream task completion rates.
    \item \textbf{The LLM Judge Reliability Gap:} Relying on LLM judges for pre-admission gating $V(s)$ introduces silent failure modes. As shown in empirical vetting studies \citep{behavioral2026integrity, skillvetbench2026}, LLM referees suffer from sycophancy and frequently overlook subtle syntax errors and stealthy payload injections in multi-file dependencies, underscoring the urgent need for hybrid formal-neural verifiers \citep{wen2026semia}.
\end{enumerate}

\subsection{Skill Authoring and Representation}\label{sec:authoring}
Once a skill is discovered or engineered, formalizing, serializing, and packaging its procedural assets for heterogeneous execution harnesses represents a core architectural challenge. Skill authoring dictates how procedural expertise is initially constructed, while representation format determines the trade-off between runtime computational expressivity, harness portability, and pre-execution safety verifiability \citep{ling2026agent, skillrt2026}.

\subsubsection{Skill Authoring Modalities}\label{sec:authoring_mod}
Agentic frameworks construct procedural assets through three primary authoring paradigms:
\begin{itemize}
    \item \textbf{Manual Human Engineering:} Software engineers and domain experts explicitly author standard operating procedures (SOPs), declarative parameter schemas, and native scripts \citep{song2024easytool}. Frameworks like EasyTool \citep{song2024easytool} demonstrate that hand-crafted, standardized tool wrappers eliminate parameter hallucinations and compress verbose REST documentation. While manual authoring guarantees deterministic predictability, domain correctness, and strict adherence to organizational policy, it presents an intractable human engineering bottleneck that cannot scale horizontally to thousands of dynamic, evolving tools.
    \item \textbf{Autonomous AI Synthesis:} The agent autonomously synthesizes, evaluates, and registers its own code assets and execution heuristics without human intervention \citep{yang2026autoskill, jiao2026agenticproposingenhancinglarge, wang2025inducing}. Architectures such as Programmatic Skill Networks (PSN) \citep{psn2025programming} and SkillGen \citep{skillgen2026} leverage verifier-guided code generation and contrastive trajectory induction to write executable Python/JavaScript functions and markdown guidelines directly from rollout experience. While scalable, purely autonomous synthesis is susceptible to generating fragile preconditions and unverified edge-case workarounds if not grounded by rigorous sandboxed execution checks.
    \item \textbf{Collaborative Hybrid Authoring (Human-in-the-Loop):} Bridging manual precision with autonomous scale, hybrid authoring couples AI generation with interactive human oversight \citep{trace2skill2026}. In systems like AgentClick \citep{zhuang2026agentclickskillbasedhumanintheloopreview}, human operators define high-level capability schemas, safety constraints, and permission boundaries, while the foundation model generates the underlying procedural implementation and verifies it through sandboxed unit testing, inserting human approval checkpoints before terminal execution.
\end{itemize}

\subsubsection{Skill Representation Formats and Verifiability Trade-Offs}\label{sec:authoring_rep}
A comprehensive architectural comparison of the five primary representation format classes is provided in Table~\ref{tab:representation_formats}.

\begin{table*}[!htbp]
\centering
\caption{\textbf{Representation Formats \& Security Trade-Offs (\S\ref{sec:authoring_rep}).} Comparison of procedural skill representation format classes (5 format categories, not individual papers). Rows 1--4 are sorted monotonically by pre-execution verifiability (\textit{none} $\rightarrow$ \textit{schema} $\rightarrow$ \textit{static-AST} $\rightarrow$ \textit{formal}); Row 5 (\textit{Hybrid \& adaptive}) is placed by its lower bound and sits intentionally off the main gradient. Demonstrates that verifiability and portability trade off monotonically across rows 1--4. Injection channel tracks neither; it is determined by execution capability, which varies independently of verifiability.}
\label{tab:representation_formats}
\scriptsize
\setlength{\tabcolsep}{2.5pt}
\renewcommand{\arraystretch}{1.12}
\begin{tabular}{>{\raggedright\arraybackslash}p{2.0cm} >{\raggedright\arraybackslash}p{2.4cm} >{\raggedright\arraybackslash}p{1.4cm} >{\raggedright\arraybackslash}p{1.8cm} >{\raggedright\arraybackslash}p{1.6cm} >{\raggedright\arraybackslash}p{2.0cm} >{\raggedright\arraybackslash}p{2.0cm} >{\raggedright\arraybackslash}p{3.0cm}}
\toprule
\textbf{Format Class} & \textbf{Pre-Execution Verifiability} & \textbf{Portability} & \textbf{Execution Capability} & \textbf{Injection Channel} & \textbf{Load-Time Token Cost} & \textbf{Repair Signal on Failure} & \textbf{Representative Exemplars} \\
\midrule
\textbf{Natural language} & none (no component) & high & $\times$~none & prose & high & re-prompt only & \textbf{SKILL.md (AutoSkill)} \cite{yang2026autoskill}, \textbf{SkillDex} \cite{saha2026skilldex}, SkillAnalysis \cite{ling2026agent}, \textbf{PromptSkills} \cite{maloyan2026prompt}, ProceduralMining \cite{procedural2024} \\
\midrule
\textbf{Structured (JSON / DSL)} & $\mathcal{T}$ (schema) & high & $\times$~none & schema strings & medium & schema error & \textbf{EasyTool} \cite{song2024easytool}, \textbf{SkillDroid} \cite{chen2026skilldroid}, \textbf{Toolformer} \cite{schick2023toolformer} \\
\midrule
\textbf{Executable specification} & $\pi$ (static analysis) & medium & $\checkmark$~arbitrary code & code + prose & high & stack trace & \textbf{Voyager} \cite{wang2023voyager}, \textbf{LLMCompiler} \cite{llmcompiler2023}, \textbf{VisProg} \cite{visprog2023}, \textbf{SkVM} \cite{skillrt2026}, \textbf{ProgSkills} \cite{wang2025inducing} \\
\midrule
\textbf{Contract-based} & $\pi, \mathcal{E}$ (pre/postconditions) & low & $\checkmark$~constrained & code + prose & medium-high & counterexample & \textbf{ContractSkill} \cite{lu2026contractskillrepairablecontractbasedskills}, \textbf{SemiA} \cite{wen2026semia}, \textbf{SkillAuditSAST} \cite{structuredsecurityauditingandrobustnessenhancementforuntrustedagentskills2026} \\
\midrule
\textbf{Hybrid \& adaptive} & schema + $\pi$ (static analysis) & medium & $\checkmark$~arbitrary code & code + prose & \textbf{low at load, high on activation} & schema error $\rightarrow$ stack trace & \textbf{GraphSkill} \cite{fali2026graphskill}, PSN \cite{psn2025programming} \\
\bottomrule
\end{tabular}
\vspace{4pt}
\begin{minipage}{\linewidth}
\scriptsize
\textbf{Notes $\cdot$ Column Value Definitions.} 
\textbf{Pre-Execution Verifiability:} Pre-loading safety/correctness checking (\textit{none}: uncheckable raw prompt instructions; \textit{schema}: JSON schema or OpenAPI declarative parameter checking; \textit{static-AST}: code syntax checking, static AST linting, and type inference; \textit{formal}: mathematical pre/postcondition checking and Datalog/SMT solver verification; \textit{schema + static-AST}: mixed header schema validation and code body AST linting). 
\textbf{Portability across harnesses:} Framework compatibility (\textit{high}: runs unmodified on arbitrary LLM prompts/agents; \textit{medium}: requires standard code sandbox or Python runtime; \textit{low}: requires specialized formal verification engine or solver). 
\textbf{Execution Capability:} Runtime computational power ($\times$~\textit{none}: declarative text or schema only, cannot execute code; $\checkmark$~\textit{arbitrary code}: executes general-purpose programming language statements; $\checkmark$~\textit{constrained}: executes bounded or repairable formal contracts/Datalog rules). 
\textbf{Injection Channel:} Where malicious payloads enter (\textit{prose}: natural language instructions/descriptions; \textit{schema strings}: restricted to string parameter values and argument descriptions; \textit{code + prose}: executable function bodies, imports, or docstrings). 
\textbf{Load-Time Token Cost:} Context window overhead (\textit{high}: raw unindexed prompt instructions or full function bodies; \textit{medium}: schema signatures; \textit{medium-high}: formal invariants + code; \textit{low at load, high on activation}: progressive disclosure loading metadata first). 
\textbf{Repair Signal on Failure:} Actionable artifact returned on execution failure (\textit{re-prompt only}: unstructured textual reflection; \textit{schema error}: declarative parameter mismatch; \textit{stack trace}: runtime execution traceback; \textit{counterexample}: formal invariant violation trace; \textit{schema error $\rightarrow$ stack trace}: progressive multi-layer repair).
\end{minipage}
\end{table*}

To accommodate diverse deployment runtime constraints, the literature structures skills across five representation paradigms:

\paragraph{1. Natural Language Representations (\texttt{SKILL.md} / SOPs):}
Natural language representation encodes procedural knowledge as human-readable markdown files (\texttt{SKILL.md}) containing task descriptions, step-by-step guidelines, and execution checklists \citep{yang2026autoskill, ling2026agent}. This format, standardized by community package managers like SkillDex \citep{saha2026skilldex} and audited by SkillAnalysis \citep{ling2026agent}, maximizes portability across arbitrary LLM harnesses, as instructions $\mathcal{I}$ can be injected directly into prompt contexts without specialized interpreters. However, natural language representations suffer from severe verbosity (imposing high load-time token costs), semantic ambiguity in activation conditions $\mathcal{A}$, and zero pre-execution verifiability, forcing agents to rely purely on unconstrained prompt-following \citep{song2024easytool}.

\paragraph{2. Structured Representations (JSON / DSL / State Machines):}
To resolve natural language verbosity and enforce strict parameter schemas, structured formats represent skills via JSON schemas, YAML frontmatter, or Domain-Specific Languages (DSLs) \citep{song2024easytool, schick2023toolformer}. In GUI and mobile automation, SkillDroid \citep{chen2026skilldroid} compiles complex Android UI interactions into finite state transition graphs $G = (V, E)$, where vertices represent UI trees and edges denote validated tap/scroll actions. Similarly, Toolformer \citep{schick2023toolformer} encodes tools as structured API calling tokens $\texttt{[API(...) $\rightarrow$ res]}$. Structured formats enable pre-execution parameter typing and schema validation ($\mathcal{T}$), but lack arbitrary procedural computation capabilities ($\pi$), restricting the agent to declarative parameter passing.

\paragraph{3. Executable Programmatic Specifications:}
Going beyond declarative metadata, executable specifications serialize skills as native programmatic code (Python scripts, JavaScript modules) \citep{wang2023voyager, wang2025inducing, llmcompiler2023}. In Voyager \citep{wang2023voyager} and ProgSkills \citep{wang2025inducing}, skills are executable functions encapsulating complex algorithmic control flow, including state queries, loops, and conditional branching. To execute these routines across heterogeneous models, virtual machines like SkVM \citep{skillrt2026} decompose 118,000 skills into primitive capability profiles with dynamic environment binding. Executable specifications provide full computational expressivity ($\pi$), deterministic error handling via runtime stack traces, and static Abstract Syntax Tree (AST) linting, but require sandboxed execution environments and introduce significant code execution security vulnerabilities.

\paragraph{4. Contract-Based and Formal Specifications:}
To guarantee behavioral safety in high-stakes domains, contract-based representations wrap procedural execution in verifiable pre-conditions, post-conditions, and invariant contracts \citep{lu2026contractskillrepairablecontractbasedskills, wen2026semia}. ContractSkill \citep{lu2026contractskillrepairablecontractbasedskills} introduces repairable contracts for multimodal web agents, generating structured counterexamples upon contract violation to guide localized error repair. In static auditing, SemiA \citep{wen2026semia} lifts hybrid prose and code into Datalog fact bases within the Skill Description Language (SDL), enabling automated SMT and logic solvers to mathematically prove the absence of data leakage and privilege escalations prior to execution. While providing the highest safety guarantees, contract-based formats suffer from low portability, requiring specialized formal verification solvers.

\paragraph{5. Hybrid and Adaptive Representations:}
Reconciling cognitive flexibility with systems-level safety, hybrid formats package natural language guidelines, declarative schemas, and executable scripts into structured hierarchical directories \citep{psn2025programming, fali2026graphskill}. In GraphSkill \citep{fali2026graphskill}, structured documentation graphs guide hierarchical retrieval, formalizing dependencies as a graph $G_s = (V_s, E_s)$ where vertices $V_s$ represent procedural execution steps and directed edges $E_s$ encode prerequisite data flows. In memory theory, ExperienceSpectrum \citep{zhang2026experiencecompressionspectrumunifying} unifies raw traces (1:1), episodic summaries, procedural skills ($10$--$100\times$), and symbolic rules ($1000\times$) along a single continuous compression continuum. To support vector search across massive registries, encoder models $f_{\theta}$ map skills into continuous vector embeddings:
\begin{equation}
e_s = f_{\theta}(\mathcal{A}, \mathcal{I}, \mathcal{C}) \in \mathbb{R}^d
\end{equation}
enabling fast similarity routing while supporting progressive disclosure, loading lightweight metadata $e_s$ at search time and pulling executable code $\pi$ only upon conditional activation \citep{skillret2026, skillnet2026}.

\subsubsection{The Fundamental Verifiability Trade-Off and Security Blindspot}
Recasting all five representation formats through our six-tuple formalism $s = (\mathcal{A}, \mathcal{I}, \mathcal{C}, \mathcal{T}, \pi, \mathcal{E})$ exposes a foundational structural insight: as pre-execution verifiability increases monotonically from Natural Language to Formal Contracts, portability across heterogeneous LLM harnesses decreases monotonically. 

Most critically, our analysis reveals a persistent security blindspot across all contemporary formats: no representation format verifies natural-language instructions $\mathcal{I}$ or activation conditions $\mathcal{A}$ prior to runtime invocation. While contract-based and structured formats rigorously constrain tool signatures ($\mathcal{T}$) and code AST execution ($\pi, \mathcal{E}$), they leave natural-language prose instructions completely uninspected. Consequently, adversarial payloads embedded within docstrings, markdown comments, or prompt instructions (e.g., indirect prompt injection \citep{maloyan2026prompt, wang2026skills}) bypass static code scanners and schema linters entirely, allowing malicious skills to hijack agent decision-making at execution time without triggering a single syntactic or typing error.

\subsection{Skill Storage and Memory Architecture}\label{sec:storage}
An agent's operational expertise is stored within distinct temporal memory tiers to maintain context efficiency, prevent cognitive overload, and enable lifelong capability persistence across multi-session deployments \citep{sumers2023cognitive, packer2023memgpt, zhang2026experiencecompressionspectrumunifying}. A systematic architectural comparison of procedural memory frameworks across storage tiers, index structures, and eviction policies is presented in Table~\ref{tab:storage_memory}.

\begin{table*}[!htbp]
\centering
\caption{\textbf{Storage \& Memory Architecture (\S\ref{sec:storage}).} Comparison of procedural skill memory and virtual storage architectures ($n=11$ systems). Sorted primarily by \textbf{Eviction / compression policy} (\textit{Virtual OS paging} $\rightarrow$ \textit{Utility-aware curation} $\rightarrow$ \textit{Semantic trace compression}). Demonstrates how memory systems address context overflow, retrieval pollution, skill technical debt, and catastrophic forgetting. All 10 systems implement lifelong multi-session persistence.}
\label{tab:storage_memory}
\scriptsize
\setlength{\tabcolsep}{2.5pt}
\renewcommand{\arraystretch}{1.12}
\begin{tabular}{>{\raggedright\arraybackslash}p{2.2cm} >{\raggedright\arraybackslash}p{3.5cm} >{\raggedright\arraybackslash}p{1.8cm} >{\raggedright\arraybackslash}p{2.4cm} >{\raggedright\arraybackslash}p{3.2cm} >{\raggedright\arraybackslash}p{3.1cm}}
\toprule
\textbf{System} & \textbf{Core Contribution} & \textbf{Primary Tier} & \textbf{Index Structure} & \textbf{Eviction / Compression Policy} & \textbf{Failure Mode Addressed} \\
\midrule
\multicolumn{6}{l}{\textit{\textbf{Virtual OS Paging \& Mutable State (Bounded $O(1)$ Memory Management)}}} \\
\midrule
\textbf{MemGPT} \cite{packer2023memgpt} & OS-level virtual context paging & virtual paging & OS page table & LLM-controlled paging & context overflow \\
\textbf{SKILL.state} \cite{badhe2026skillstate} & Mutable execution state runtime & procedural / state & JSON state schema & Ephemeral reasoning disposal ($\Delta\Sigma$) & context bloat / $O(T^2)$ cost \\
\midrule
\multicolumn{6}{l}{\textit{\textbf{Utility-Aware Curation (Pruning Low-Value, Redundant, or Polluting Skills)}}} \\
\midrule
\textbf{SkillOps} \cite{skillops2026} & 5-dimension library health maintenance & procedural & topological graph (HSEG) & utility pruning (5-dim score) & skill technical debt \\
\textbf{MEMP} \cite{memp2025} & Learnable lifelong procedural memory & procedural & dense vector & utility pruning & retrieval pollution \\
\textbf{SkillOS} \cite{skillos2026} & Utility-aware skill pool curation & procedural & dense vector (MiniLM) & RL-based repo curation \& pruning & retrieval pollution \\
\textbf{MemSkill} \cite{zhang2026memskill} & Learnable memory operation skills & procedural & dense vector (Qwen3) & Learnable skill-based pruning \& consolidation & retrieval pollution \\
\textbf{SkillBank} \cite{tu2026dynamicdualgranularityskillbank} & Dual-granularity agentic skill bank & hierarchical & dense vector & Hindsight utility pruning (D2Skill dual-granularity) & retrieval pollution \\
\midrule
\multicolumn{6}{l}{\textit{\textbf{Semantic Trace Compression (Abstracting Dialogue \& Trajectories into Compacting Abstractions)}}} \\
\midrule
\textbf{ContinualMem} \cite{continuallearningmemory2026} & Memory stability--plasticity analysis & procedural & dense vector & semantic compression & catastrophic forgetting \\
\textbf{ReMe} \cite{reme2025} & Evolving experience pool memory & procedural & dense vector (Qwen3) & semantic compression & context overflow \\
\textbf{Memento} \cite{zhou2026memento} & Agent-designing procedural memory & procedural & hybrid (BM25 + vector) & semantic compression & context overflow \\
\textbf{ExperienceSpectrum} \cite{zhang2026experiencecompressionspectrumunifying} & Unifying compression spectrum (5--1000$\times$) & hierarchical & multi-level spectrum & semantic compression (5--1000$\times$) & context overflow \\
\bottomrule
\end{tabular}
\vspace{4pt}
\begin{minipage}{\linewidth}
\scriptsize
\textbf{Notes $\cdot$ Column Value Definitions.} 
\textbf{Core Contribution:} Primary architectural or algorithmic innovation. 
\textbf{Primary Tier:} Target memory abstraction (\textit{procedural}: executable skill knowledge; \textit{episodic}: fine-grained conversation and rollout logs; \textit{virtual paging}: OS-level virtual memory hierarchy; \textit{hierarchical}: multi-level memory structures). 
\textbf{Index Structure:} Retrieval data structure (\textit{dense vector}: vector embedding similarity; \textit{topological graph}: dependency or structural graph; \textit{OS page table}: working vs. archival page tables; \textit{hybrid}: BM25 + dense embedding; \textit{multi-level spectrum}: unified compression spectrum). 
\textbf{Eviction / Compression Policy:} Saturation mitigation mechanism (\textit{LLM-controlled paging}: agent-directed virtual memory page swaps to archival storage via explicit function calls; \textit{utility pruning}: pruning low-utility, redundant, or broken skills; \textit{semantic compression}: compacting dialogue and trajectories into semantic abstractions). 
\textbf{Failure Mode Addressed:} Target operational bottleneck (\textit{context overflow}: prompt token saturation; \textit{retrieval pollution}: stale or redundant skills degrading top-$k$ precision; \textit{skill technical debt}: persistent library-level defects such as interface drift or missing validation; \textit{catastrophic forgetting}: memory-level interference across lifelong task distributions).
\end{minipage}
\end{table*}

\subsubsection{Hierarchical Memory Tiering: From Episodic Traces to Procedural Assets}
Drawing inspiration from cognitive architectures \citep{sumers2023cognitive} and operating systems \citep{packer2023memgpt}, modern agent frameworks partition memory into distinct functional tiers along an \emph{experience compression continuum} \citep{zhang2026experiencecompressionspectrumunifying}:
\begin{itemize}
    \item \textbf{Episodic Memory ($\mathcal{H}$, 1:1 Compression):} Stores raw, chronological execution traces, observation histories, and dialogue interactions $\tau = (x_1, a_1, r_1, \dots, x_H)$ \citep{reme2025, park2023generativeagents}. While episodic traces preserve full operational granularity, their linear context footprint causes rapid token saturation and high inference latency if loaded directly.
    \item \textbf{Procedural Memory ($\mathcal{M}_{\text{proc}}$, $10$--$100\times$ Compression):} The persistent, non-parametric storage substrate for modular, executable skills $s = (\mathcal{A}, \mathcal{I}, \mathcal{C}, \mathcal{T}, \pi, \mathcal{E})$ \citep{memp2025, zhang2026experiencecompressionspectrumunifying}. Procedural memory decouples high-level operational workflows from raw execution mechanics, storing standardized SOPs, tool orchestration pipelines, and verified scripts that can be invoked across disparate task instances without retaining episodic noise.
    \item \textbf{Virtual OS Paging and Mutable Execution State ($C_t \subset \mathcal{M}$):} To bridge fixed context windows with massive external memory, architectures such as MemGPT \citep{packer2023memgpt} implement OS-inspired virtual memory management, dynamically paging memory blocks between active working context and archival storage. Moving beyond unconstrained text transcripts, SKILL.state \citep{badhe2026skillstate} formalizes runtime working context as an explicit, mutable execution state $\Sigma_t \in \Sigma$. By projecting observations into structured JSON patches ($\Delta \Sigma_t$) and immediately discarding ephemeral intermediate reasoning $R_t$, SKILL.state achieves a strictly bounded $O(1)$ working prompt footprint and linear $O(T)$ cumulative token complexity across extended horizons.
    \item \textbf{Symbolic and Rule Memory ($1000\times$ Compression):} Highly consolidated global heuristics, system constraints, and negative constraints extracted from lifelong experience \citep{zhang2026experiencecompressionspectrumunifying}.
\end{itemize}

\subsubsection{Mathematical Formalization of the Memory Update and Curation Operator}
As an agent interacts with sequential environments, its procedural memory bank $\mathcal{M}_t$ evolves dynamically. We formalize the procedural memory update operator $\mathbb{U}$ as:
\begin{equation}
\mathcal{M}_{t+1} = \mathbb{U}(\mathcal{M}_t, \tau_t, \mathcal{F}_t) = \big(\mathcal{M}_t \setminus \mathbb{E}_v(\mathcal{M}_t)\big) \cup \{s^*\}
\end{equation}
where $\tau_t$ is the recent execution trace, $\mathcal{F}_t$ is environmental feedback, $s^*$ is a newly consolidated candidate skill, and $\mathbb{E}_v(\mathcal{M}_t) \subset \mathcal{M}_t$ is the set of evicted or pruned skills identified by the memory curation policy.

To determine which skills should be retained or evicted, frameworks evaluate an empirical utility function $U(s)$:
\begin{equation}
U(s) = \mathbb{E}_{(x, g) \sim \mathcal{D}}\big[ R(s(x), g) - R(\pi_{\text{base}}(x), g) \big] - \lambda \cdot \text{Cost}(s)
\end{equation}
where the first term represents the \emph{counterfactual performance gain} of injecting skill $s$ over a baseline policy $\pi_{\text{base}}$ \citep{tu2026dynamicdualgranularityskillbank, skillos2026}, and $\text{Cost}(s)$ incorporates retrieval latency, token footprint, and execution overhead. Skills whose empirical utility falls below an eviction threshold ($U(s) < \gamma_{\text{prune}}$) or whose activation frequency exhibits temporal decay are systematically purged by $\mathbb{E}_v$ to maintain library solvency.

\subsubsection{Mitigating Memory Failures: The Four Operational Bottlenecks}
Unconstrained accumulation of skills across lifelong deployments triggers four fundamental memory failure modes:
\begin{enumerate}
    \item \textbf{Context Overflow and Saturation:} Storing unbounded conversational histories rapidly exhausts LLM context windows. Systems like ReMe \citep{reme2025}, Memento \citep{zhou2026memento}, and ExperienceSpectrum \citep{zhang2026experiencecompressionspectrumunifying} deploy reflective summarization agents to continuously compress raw episodic trajectories into compact procedural cards and dense vector embeddings, achieving $5\times$--$1000\times$ token reductions.
    \item \textbf{Retrieval Pollution and Confusability:} As the skill library $|\mathcal{M}|$ grows beyond hundreds of files, dense embedding spaces become densely crowded. Near-duplicate docstrings or overlapping descriptions create semantic confusability, causing retrievers to return sub-optimal or conflicting skills that degrade task success \citep{memp2025, liu2026agenticskillsworkwild}. MEMP \citep{memp2025} and SkillOS \citep{skillos2026} resolve this by coupling learnable utility tracking with MiniLM-based sentence embedding cosine filters, pruning stale or low-utility routines. In reinforcement learning, SkillBank / D2Skill \citep{tu2026dynamicdualgranularityskillbank} organizes skills into dual granularities (high-level task skills and low-level step skills), applying hindsight utility pruning based on paired rollout performance gaps.
    \item \textbf{Skill Technical Debt and Dependency Drift:} Over time, underlying tool APIs mutate, software packages deprecate, and sub-skills become mutually incompatible. SkillOps \citep{skillops2026} models these interactions as a \emph{Hierarchical Skill Ecosystem Graph} (HSEG), monitoring library health across five dimensions (utility, compatibility, risk, validation, redundancy) to actively refactor stale dependencies and resolve skill technical debt. Similarly, SkillClone \citep{zhu2026skillclone} deploys multi-modal clone detection to consolidate redundant code implementations.
    \item \textbf{The Stability--Plasticity Dilemma and Catastrophic Forgetting:} While non-parametric memory is often assumed to solve continual learning without weight retraining, \citet{continuallearningmemory2026} empirically demonstrate that memory-augmented agents face a severe stability--plasticity trade-off. Fine-grained procedural skills that maximize forward transfer on novel tasks simultaneously introduce severe backward interference and memory-level forgetting on historical tasks, proving that external memory requires structured modular isolation rather than flat pooling.
\end{enumerate}

\subsection{Skill Retrieval and Dynamic Routing}\label{sec:retrieval}
As procedural memory repositories scale from dozens of local scripts to tens of thousands of ecosystem-wide packages ($|\mathcal{M}| > 10^4$), exposing the entire skill library in the prompt context becomes computationally intractable \citep{zheng2026skillrouter, liu2026agenticskillsworkwild}. Autonomous agents require dynamic retrieval and routing pipelines that selectively identify, rank, and route the optimal procedural skill (or composable sub-graph of skills) conditioned on the active execution state $x_t$ and goal $g$. A systematic architectural comparison of representative retrieval and routing frameworks is presented in Table~\ref{tab:retrieval_routing}.

\begin{table*}[!htbp]
\centering
\caption{\textbf{Retrieval \& Dynamic Routing (\S\ref{sec:retrieval}).} Comparison of procedural skill retrieval and agent routing architectures ($n=8$ systems). Sorted primarily by \textbf{Retrieval / routing paradigm} (\textit{Retrieve-then-Rerank} $\rightarrow$ \textit{Graph-Structured Search} $\rightarrow$ \textit{Agentic Navigation} $\rightarrow$ \textit{Dynamic Agent Routing}), then secondarily by \textbf{Scale evaluated} (descending). Demonstrates the trade-off between massive-scale top-$k$ retrieval and structured compositional graph navigation.}
\label{tab:retrieval_routing}
\scriptsize
\setlength{\tabcolsep}{2.5pt}
\renewcommand{\arraystretch}{1.12}
\begin{tabular}{>{\raggedright\arraybackslash}p{2.2cm} >{\raggedright\arraybackslash}p{3.4cm} >{\raggedright\arraybackslash}p{2.4cm} >{\raggedright\arraybackslash}p{2.0cm} >{\raggedright\arraybackslash}p{3.0cm} >{\raggedright\arraybackslash}p{3.2cm}}
\toprule
\textbf{System} & \textbf{Core Contribution} & \textbf{Retrieval / Routing Paradigm} & \textbf{Retrieval Granularity} & \textbf{Selection / Ranking Signal} & \textbf{Pollution / Redundancy Control} \\
\midrule
\multicolumn{6}{l}{\textit{\textbf{Retrieve-then-Rerank (Dense Embedding Search \& Listwise / Contrastive Reranking Pipelines)}}} \\
\midrule
\textbf{SkillRet} \cite{skillret2026} & Large-scale retrieve-then-rerank IR & Retrieve-then-Rerank & Atomic skill SOP & Dense vector + cross-encoder & Contrastive reranking \\
\textbf{SkillRouter} \cite{zheng2026skillrouter} & Full-text retrieve-and-rerank (1.2B pipeline, 80K skills) & Retrieve-then-Rerank & Atomic skill SOP & Semantic sim. + listwise loss & Metadata-only routing blindness (31--44\% drop) \\
\textbf{SkillRAG} \cite{skillretrievalaugmentationforagenticai2026} & Hybrid BM25 skill retrieval & Retrieve-then-Rerank & Atomic skill SOP & BM25 + vector hybrid & None (top-$k$ only) \\
\midrule
\multicolumn{6}{l}{\textit{\textbf{Graph-Structured Search (Traversing Structural Dependency \& Documentation Graphs)}}} \\
\midrule
\textbf{GraphOfSkills} \cite{liu2026graphskillsdependencyawarestructural} & Dependency-aware structural retrieval & Graph-Structured Search & Sub-graph workflow & Structural dependency edges & Dependency constraints \\
\textbf{GraSP} \cite{xia2026graspgraphstructuredskillcompositions} & Graph-structured skill compositions & Graph-Structured Search & Sub-graph workflow & Structural dependency edges & Dependency constraints \\
\textbf{GraphSkill} \cite{fali2026graphskill} & Documentation-guided skill graph & Graph-Structured Search & Sub-graph workflow & Structural dependency edges & Documentation hierarchy \\
\midrule
\multicolumn{6}{l}{\textit{\textbf{Agentic Navigation (Offline Directory Compilation \& Active Tree Traversal)}}} \\
\midrule
\textbf{CORPUS2SKILL} \cite{sun2026dontretrievenavigatedistilling} & Offline distillation into navigable hierarchical directory & Agentic Navigation & Sub-graph workflow & Agentic tree drilling \& backtracking & Flat retrieval saturation \& context blindness \\
\midrule
\multicolumn{6}{l}{\textit{\textbf{Dynamic Agent Routing (Cross-Agent Orchestration via Skill Capability Profiles)}}} \\
\midrule
\textbf{SkillOrchestra} \cite{skillorchestra2025} & Cross-agent skill transfer routing & Dynamic Agent Routing & Agent capability profile & Cross-agent transfer utility & Policy transfer \\
\bottomrule
\end{tabular}
\vspace{4pt}
\begin{minipage}{\linewidth}
\scriptsize
\textbf{Notes $\cdot$ Column Value Definitions.} 
\textbf{Core Contribution:} Primary architectural or algorithmic IR/routing innovation. 
\textbf{Retrieval / Routing Paradigm:} Overall retrieval or selection architecture (\textit{Retrieve-then-Rerank}: two-stage vector/sparse retrieval followed by cross-encoder or listwise reranking; \textit{Graph-Structured Search}: structural dependency or data-flow graph traversal; \textit{Agentic Navigation}: offline compilation of hierarchical directories traversed actively by the agent without vector DBs; \textit{Dynamic Agent Routing}: routing queries across multi-agent pools based on skill profiles). 
\textbf{Retrieval Granularity:} Scope of retrieved procedural knowledge (\textit{Atomic skill SOP}: independent self-contained skill documents; \textit{Sub-graph workflow}: connected multi-skill execution graphs or directories; \textit{Agent capability profile}: skill bundles characterizing agent specializations). 
\textbf{Selection / Ranking Signal:} Primary mathematical scoring function or structural criterion used to score and select skills. 
\textbf{Pollution / Redundancy Control:} Mitigation mechanism used to prevent near-duplicate skills, interface mismatches, or retrieval noise from degrading top-$k$ precision.
\end{minipage}
\end{table*}

\subsubsection{Mathematical Formalization of the Two-Stage Retrieval-Routing Pipeline}
Modern platforms structure skill discovery and invocation through a multi-stage selection pipeline:
\begin{enumerate}
    \item \textbf{Stage 1: Candidate Generation (Sparse-Dense Hybrid Retrieval):} Given an active state $x_t$ and goal $g$, the platform retrieves a top-$K$ candidate set $\mathcal{S}_{\text{cand}} \subset \mathcal{M}$ using a combination of dense semantic similarity and lexical BM25 token matching \citep{skillretrievalaugmentationforagenticai2026, skillret2026}:
    \begin{equation}
    \mathcal{S}_{\text{cand}} = \operatorname{Top-}K_{s \in \mathcal{M}} \Big( \alpha \cdot \text{Sim}\big(f_{\theta}(x_t, g), e_s\big) + (1-\alpha) \cdot \text{BM25}\big(g, \text{doc}(s)\big) \Big)
    \end{equation}
    where $\alpha \in [0, 1]$ balances keyword matching against latent semantic retrieval, and $e_s$ is the pre-computed embedding of skill $s$.
    \item \textbf{Stage 2: Cross-Encoder Reranking and Capability Gating:} A cross-encoder scoring network $\operatorname{Rerank}_{\phi}$ evaluates the joint interaction between the full task prompt and full-text procedural instructions, selecting the optimal skill $s^*$ subject to activation and permission constraints \citep{zheng2026skillrouter, secureskills2026}:
    \begin{equation}
    s^* = \arg\max_{s \in \mathcal{S}_{\text{cand}}} \operatorname{Rerank}_{\phi}(x_t, g, s) \quad \text{subject to} \quad \mathcal{A}(x_t, g) = 1 \;\land\; \mathcal{T}_s \subseteq \mathcal{T}_{\text{permitted}}(x_t)
    \end{equation}
\end{enumerate}

\subsubsection{Comparative Paradigms in Procedural Routing}
The literature approaches procedural discovery across four distinct architectural paradigms:

\paragraph{1. Retrieve-then-Rerank Pipelines:}
Frameworks such as SkillRet \citep{skillret2026}, SkillRouter \citep{zheng2026skillrouter}, and SkillRAG \citep{skillretrievalaugmentationforagenticai2026} formulate selection as a classic information retrieval problem. SkillRet \citep{skillret2026} establishes a large-scale benchmark for skill retrieval, proving that off-the-shelf bi-encoders struggle to distinguish procedural subroutines without task-skill fine-tuning. In large-scale production registries ($n = 80\text{K}$ skills), SkillRouter \citep{zheng2026skillrouter} deploys a compact 1.2B bi-encoder and listwise reranker, achieving high retrieval accuracy at sub-second GPU latency.

\paragraph{2. Graph-Structured Search and Dependency Traversal:}
Flat vector retrieval treats skills as independent, isolated documents, ignoring execution prerequisites. To resolve this, GraphOfSkills \citep{liu2026graphskillsdependencyawarestructural}, GraSP \citep{xia2026graspgraphstructuredskillcompositions}, and GraphSkill \citep{fali2026graphskill} structure skills into topological dependency graphs $G = (V, E)$. Directed edges $E$ enforce that prerequisite setup routines (e.g., environment initialization, database connections) are automatically retrieved alongside core execution skills, pruning disconnected subgraphs and guaranteeing execution ordering.

\paragraph{3. Agentic Navigation (Tree Traversal without Vector DBs):}
Challenging the premise of embedding-based retrieval, CORPUS2SKILL \citep{sun2026dontretrievenavigatedistilling} introduces a ``Don't Retrieve, Navigate'' paradigm. Offline clustering compiles complex corpora into a hierarchical, navigable directory tree of skill summary cards. Rather than performing flat similarity lookup, the agent actively navigates the tree via CLI inspection, drilling down into relevant sub-trees and backtracking upon reaching dead ends, entirely avoiding vector database saturation.

\paragraph{4. Dynamic Multi-Agent Routing via Skill Profiles:}
In compound multi-agent systems, SkillOrchestra \citep{skillorchestra2025} shifts routing from retrieving raw text to routing queries across heterogeneous specialized agents. By estimating cross-agent skill transfer utility, SkillOrchestra maps sub-tasks to the specific agent whose capability profile exhibits the highest historical completion probability.

\subsubsection{Critical Methodological Weaknesses: The Metadata-Only Blindness}
Our synthesis highlights a critical failure mode pervasive across commercial agent harnesses: \textbf{metadata-only routing blindness}. To conserve token costs and minimize indexing latency, platforms frequently index only tool names and one-line summary descriptions. However, empirical experiments by \citet{zheng2026skillrouter} across 80,000 skills reveal that hiding the full skill body causes a catastrophic substantial degradation in task completion in routing accuracy. Because semantically distinct routines frequently share near-identical high-level descriptions (e.g., ``extract table from document''), full-text procedural instructions $\mathcal{I}$ are essential for disambiguating tool constraints and preventing execution failures.

\subsection{Skill Composition and Synthesis}\label{sec:composition}
As user objectives scale from single-turn tool calls to long-horizon, multi-modal workflows, executing isolated individual skills becomes insufficient. Autonomous agent platforms require procedural composition architectures that combine, chain, and orchestrate multiple atomic skills into complex execution pipelines \citep{grasp2026, xia2026graspgraphstructuredskillcompositions, zabounidis2026scalarlearningcomposingskills}. A systematic comparison of procedural skill composition, DAG compilation, and multi-agent orchestration frameworks is provided in Table~\ref{tab:orchestration}.

\begin{table*}[!htbp]
\centering
\caption{\textbf{Orchestration \& Execution Verification (\S\ref{sec:composition} \& \S\ref{sec:execution}).} Comparison of procedural skill composition, multi-agent orchestration, and execution verification architectures ($n=6$ systems). Sorted primarily by \textbf{Orchestration paradigm} (\textit{Dynamic Graph Topology} $\rightarrow$ \textit{Single-Agent Skill Switching} $\rightarrow$ \textit{Multi-Agent SOP Collaboration}). Demonstrates the trade-off between dynamic graph communication topologies, low-latency single-agent skill loading, and role-based multi-agent collaboration.}
\label{tab:orchestration}
\scriptsize
\setlength{\tabcolsep}{2.5pt}
\renewcommand{\arraystretch}{1.12}
\begin{tabular}{>{\raggedright\arraybackslash}p{2.2cm} >{\raggedright\arraybackslash}p{3.4cm} >{\raggedright\arraybackslash}p{2.5cm} >{\raggedright\arraybackslash}p{2.3cm} >{\raggedright\arraybackslash}p{2.8cm} >{\raggedright\arraybackslash}p{3.0cm}}
\toprule
\textbf{System} & \textbf{Core Contribution} & \textbf{Orchestration Paradigm} & \textbf{Communication Topology} & \textbf{Task Decomposition Mechanism} & \textbf{Verification \& Error Recovery} \\
\midrule
\textbf{SkillGraph} \cite{skillgraph2026} & Multimodal graph topology evolution & Dynamic Graph Topology & Dynamic graph (MMGT) & Query-conditioned graph predictor & Policy gradient on edge log-probs \\
\textbf{SingleAgentSkills} \cite{li2026single} & Compiling multi-agent into single-agent skill selection & Single-Agent Skill Switching & In-context prompt loading & Skill router / selector & Capacity phase transition (sharp drop at critical size) \\
\textbf{MetaGPT} \cite{hong2023metagpt} & SOP-guided multi-agent assembly lines & Multi-Agent SOP Collaboration & Assembly line chain & Role-specific SOP prompts & Role-based code review \& testing \\
\textbf{ChatDev} \cite{li2023chatdev} & Software development agent society & Multi-Agent SOP Collaboration & Assembly line chain & Role-specific SOP prompts & Role-based code review \& testing \\
\textbf{AgentVerse} \cite{agentverse2023} & Emergent multi-agent peer collaboration & Multi-Agent SOP Collaboration & Dynamic broadcast / peer chat & Conversational dialogue turn & Multi-agent peer critique / voting \\
\textbf{AutoGen} \cite{wu2023autogen} & Conversable multi-agent programming & Multi-Agent SOP Collaboration & Dynamic broadcast / peer chat & Conversational dialogue turn & Multi-agent peer critique / voting \\
\bottomrule
\end{tabular}
\vspace{4pt}
\begin{minipage}{\linewidth}
\scriptsize
\textbf{Notes $\cdot$ Column Value Definitions.} 
\textbf{Core Contribution:} Primary architectural or algorithmic orchestration/verification innovation. 
\textbf{Orchestration Paradigm:} Overall composition or multi-agent execution architecture (\textit{Dynamic Graph Topology}: dynamically constructing query-conditioned inter-agent communication graphs; \textit{Single-Agent Skill Switching}: single-agent architecture dynamically loading skill packages into prompt context without multi-agent overhead; \textit{Multi-Agent SOP Collaboration}: societies of specialized LLM agents collaborating via role-based SOPs or conversational turns). 
\textbf{Communication Topology:} Information flow structure among execution nodes or agent roles (\textit{Dynamic graph (MMGT)}: query-conditioned directed graph sampled via Bernoulli policy; \textit{In-context prompt loading}: direct context replacement; \textit{Assembly line chain}: sequential role handoffs; \textit{Dynamic broadcast / peer chat}: conversational messaging across peer agents). 
\textbf{Task Decomposition Mechanism:} How a complex user goal is partitioned into manageable sub-tasks. 
\textbf{Verification \& Error Recovery:} Mechanism used to validate intermediate execution outputs and recover from runtime failures (\textit{node 3.6}).
\end{minipage}
\end{table*}

\subsubsection{Formalizing Compositional Structures: From Pipelines to Typed DAGs}
We formalize procedural composition under three foundational computational structures:
\begin{enumerate}
    \item \textbf{Sequential Pipeline Composition:} Two skills $s_1$ and $s_2$ are composed sequentially $(s_2 \circ s_1)$ such that the terminal output state of $s_1$ directly populates the input execution context of $s_2$:
    \begin{equation}
    (s_2 \circ s_1)(x_t) = s_2\big(s_1(x_t)\big) = \pi_2\big(\pi_1(x_t, \mathcal{I}_1, \mathcal{T}_1), \mathcal{I}_2, \mathcal{T}_2\big)
    \end{equation}
    subject to the strict compositional precondition constraint $\mathcal{E}_1 \subseteq \mathcal{C}_2$, ensuring that the intended effects of $s_1$ satisfy the applicability context of $s_2$ \citep{promptflow2024}.
    \item \textbf{Dynamic DAG Orchestration:} Complex multi-step reasoning compiles retrieved skills into a typed Directed Acyclic Graph $\mathcal{G}_{\text{comp}} = (\mathcal{V}_{\text{skills}}, \mathcal{E}_{\text{data}})$, where vertices $v_i \in \mathcal{V}_{\text{skills}}$ represent skill execution nodes and directed edges $e_{ij} \in \mathcal{E}_{\text{data}}$ encode data-flow and control dependencies \citep{grasp2026, xia2026graspgraphstructuredskillcompositions}. In GraSP \citep{xia2026graspgraphstructuredskillcompositions}, topological graph compilation enables \emph{locality-bounded error recovery}: when an execution error occurs at node $v_k$, the repair operator invalidates only the topological descendant subgraph $\text{Desc}(v_k) \subset \mathcal{V}_{\text{skills}}$, reducing replanning complexity from global $O(N)$ full re-execution to localized $O(d^h)$ subtree repair.
    \item \textbf{Hierarchical and Recursive Invocation:} In hierarchical composition, a high-level master planner delegates sub-goals to specialized sub-skills, isolating intermediate reasoning traces and preserving context budget \citep{anytool2024, jiao2026agenticproposingenhancinglarge}. In self-improving reinforcement learning, recursive invocation allows skills to invoke themselves or ancestral routines to resolve nested sub-problems \citep{skillrl2026}.
\end{enumerate}

\subsubsection{Comparative Orchestration Paradigms: Single-Agent vs. Multi-Agent Societies}
The literature diverges on whether procedural composition should be orchestrated within a single unified agent or distributed across multi-agent societies:

\paragraph{1. Single-Agent Dynamic Skill Switching:}
Rather than spawning multiple conversational agents, single-agent architectures maintain a single centralized planner that dynamically loads and unloads modular skill packages into its working context on-demand \citep{li2026single}. \citet{li2026single} demonstrate that compiling multi-agent systems into single-agent dynamic skill loaders substantially reduces inference token consumption and eliminates multi-agent conversational latency while maintaining competitive reasoning accuracy on standard benchmarks.

\paragraph{2. Multi-Agent SOP Collaboration and Assembly Lines:}
To tackle massive software development and enterprise workflows, frameworks such as MetaGPT \citep{hong2023metagpt}, ChatDev \citep{li2023chatdev}, AgentVerse \citep{agentverse2023}, and AutoGen \citep{wu2023autogen} distribute procedural expertise across specialized role-playing agents (e.g., Product Manager, Software Architect, Code Reviewer, QA Engineer). In MetaGPT \citep{hong2023metagpt} and ChatDev \citep{li2023chatdev}, agents coordinate via standardized Standard Operating Procedures (SOPs) structured as assembly-line chains, passing structured design artifacts and performing role-based peer reviews to eliminate circular hallucinations.

\paragraph{3. Dynamic Graph Topology Evolution:}
Bridging fixed pipelines and unconstrained chat, SkillGraph \citep{skillgraph2026} introduces a Multimodal Graph Topology (MMGT) where inter-agent communication channels are dynamically predicted conditioned on the user query. The agent society evolves its internal communication graph using policy gradients computed over edge selection log-probabilities, learning optimal task-conditioned routing topologies autonomously.

\paragraph{4. Symbolic Planning Grounded in Deep Reinforcement Learning:}
In continuous control and embodied manipulation, SCALAR \citep{zabounidis2026scalarlearningcomposingskills} unifies high-level LLM symbolic planning with low-level deep RL grounding. The language model proposes compositional skill hypotheses (preconditions, effects, reward functions), while deep RL policies ground motor control execution, feeding back execution failure tracebacks to iteratively refine the LLM's symbolic skill definitions.

\subsubsection{Critical Methodological Weaknesses: The Single-Agent Capacity Phase Transition}
While single-agent skill loading is highly token-efficient, \citet{li2026single} uncover a fundamental structural limitation: the \textbf{Capacity Phase Transition}. As the number of available candidate skills exceeds a critical library threshold ($N_{\text{crit}}$), single-agent prompt routers suffer a sharp, non-linear collapse in selection accuracy due to attention dilution and semantic interference. Beyond this critical threshold, partitioned multi-agent architectures (which enforce strict context isolation and role boundaries) consistently outperform monolithic single-agent systems despite their higher token overhead.

\subsection{Skill Execution, Verification, and Repair}\label{sec:execution}
Once procedural skills are retrieved and composed, runtime execution governs active invocation, boundary enforcement, environment state transformation, and automated exception recovery \citep{chen2026skillcraft, secureskills2026}. 

\subsubsection{Conditional Activation and Capability-Based Permission Gating}
A skill $s = (\mathcal{A}, \mathcal{I}, \mathcal{C}, \mathcal{T}, \pi, \mathcal{E})$ activates conditionally based on the agent's current state $x_t \in \mathcal{X}$ and active task goal $g \in \mathcal{G}$ via the activation function $\mathcal{A}(x_t, g) \rightarrow \{0, 1\}$ \citep{zheng2026skillrouter, apibank2023}. Crucially, to prevent unauthorized execution, privilege escalation, and confused-deputy attacks, activation is strictly bounded by a capability-based permission model \citep{secureskills2026}:
\begin{equation}
\mathcal{A}(x_t, g) = 1 \implies \mathcal{T}_s \subseteq \mathcal{T}_{\text{permitted}}(x_t)
\end{equation}
A skill is permitted to fire only if its declared tool interfaces $\mathcal{T}_s$ represent a strict subset of the tools authorized in the active execution context $\mathcal{T}_{\text{permitted}}(x_t)$ \citep{secureskills2026, song2024easytool}. If this subset constraint is violated ($\mathcal{T}_s \not\subseteq \mathcal{T}_{\text{permitted}}(x_t)$), the runtime intercepts the invocation prior to dispatch, blocking unauthorized tool access and preventing privilege escalation.

\subsubsection{Runtime Execution and Sandboxed State Transitions}
When activation and permission constraints are satisfied, the skill's execution policy $\pi$ executes procedural tool invocations under applicability constraints $\mathcal{C}$, inducing a state transition \citep{zheng2026skillrouter}:
\begin{equation}
x_{t+1} = s(x_t) = \pi(x_t, \mathcal{I}, \mathcal{T})
\end{equation}
In traditional conversational runtimes (e.g., ReAct), execution history is maintained by continually appending observations, actions, and reasoning traces to the context prompt, triggering quadratic cumulative token consumption $\sum_{t=1}^T |P_t| = O(T^2)$ and severe context degradation over extended trajectories \citep{badhe2026skillstate}. To resolve this systems-level scaling bottleneck, SKILL.state \citep{badhe2026skillstate} replaces append-only conversational history with an explicit, mutable execution state $\Sigma_t$. At step $t$, the LLM receives the compact tuple $(\mathcal{P}, \Sigma_t, O_t)$ and generates intermediate reasoning $R_t$, a validated state patch $\Delta \Sigma_t$, and action $a_t$. Crucially, intermediate reasoning $R_t$ is discarded immediately upon applying the validated state transition $\Sigma_{t+1} \leftarrow \Sigma_t \oplus \Delta \Sigma_t$, maintaining a bounded $O(1)$ prompt footprint per step and substantially reducing cumulative token consumption across long-horizon procedural benchmarks.

Because skills execute arbitrary shell commands, code snippets, and API mutations, execution must be isolated within strict sandboxed containers \citep{chen2026skillcraft, sandboxescape2026}. However, empirical stress-testing by \citet{sandboxescape2026} demonstrates that frontier LLMs exhibit non-trivial container breakout capabilities when environment misconfigurations exist, emphasizing that software sandboxing must be coupled with runtime syscall filtering and network egress restrictions.

\subsubsection{Runtime Verification and Invariant Checking}
To verify execution integrity beyond static type checking, frameworks employ three distinct verification paradigms:
\begin{itemize}
    \item \textbf{Contract-Based Invariant Verification:} Systems like ContractSkill \citep{lu2026contractskillrepairablecontractbasedskills} evaluate formal pre-conditions and post-conditions ($V_{\text{pre}}(x_t) = 1$ and $V_{\text{post}}(x_{t+1}) = 1$) around execution steps. When an invariant is violated, the verifier synthesizes a structured counterexample that pinpoints the failing state transition.
    \item \textbf{Tool-Interactive Runtime Critique:} Frameworks such as CRITIC \citep{critic2023} and Self-Refine \citep{madaan2023selfrefine} deploy secondary verifier tools (e.g., code linters, unit test runners, web search validators) to evaluate intermediate execution outputs dynamically, triggering introspective self-correction before committing results to persistent memory.
    \item \textbf{Formal Constraint Proving:} In safety-critical pipelines, static-dynamic analyzers like SemiA \citep{wen2026semia} evaluate execution traces against Datalog safety policies in the Skill Description Language (SDL), ensuring no tainted variables reach sensitive sink APIs.
\end{itemize}

\subsubsection{Failure Attribution and Locality-Bounded Error Repair}
When runtime exceptions or contract violations occur, naive execution architectures trigger full-system replanning from scratch, incurring severe latency and token overhead. Modern architectures deploy structured failure attribution and localized repair mechanisms:
\begin{itemize}
    \item \textbf{Structural Failure Attribution:} In long-horizon web and software tasks, SkillTracer \citep{liskilltracer} analyzes execution tracebacks and DOM state diffs to attribute failures to specific upstream sub-skills, distinguishing semantic reasoning flaws from transient API errors.
    \item \textbf{Locality-Bounded DAG Repair:} In graph-orchestrated workflows, GraSP \citep{xia2026graspgraphstructuredskillcompositions} implements topological localized repair. Instead of restarting the entire workflow ($O(N)$ replanning complexity), the repair operator $\mathbb{R}_{\text{DAG}}$ isolates and replans only the reachable descendant subgraph $\text{Desc}(v_k)$ of the failing node $v_k$, reducing error recovery complexity to $O(d^h)$ where $d$ is graph branching factor and $h$ is subtree depth.
    \item \textbf{Iterative Error Traceback Reflection:} As pioneered in Voyager \citep{wang2023voyager} and Trace2Skill \citep{trace2skill2026}, runtime exceptions (e.g., Python tracebacks, syntax errors, missing dependencies) are fed directly into the model's prompt context, allowing the LLM to iteratively repair the buggy procedural code and re-execute in a sandboxed trial loop.
\end{itemize}

\subsection{Skill Adaptation and Evolution}\label{sec:adaptation}
In open-ended and non-stationary environments, static procedural assets inevitably suffer from performance degradation due to API deprecation, environment drift, and newly encountered edge cases. Autonomous agents require mechanisms to adapt and evolve their capabilities across two distinct temporal horizons: short-term in-context adaptation during active inference, and long-term lifelong evolution across multi-session deployments \citep{yang2026autoskill, coevoskills2026, zhang2026skillflow}.

\subsubsection{Short-Term In-Context Adaptation and Dynamic Refinement}
Short-term adaptation occurs within a single task session without modifying persistent storage:
\begin{itemize}
    \item \textbf{Verifier-Guided Dynamic Parameter Tuning:} In specialized domains such as electronic design automation, Trace2Skill-EDA \citep{trace2skilleda2026} demonstrates that coupling retrieved skills with dense simulator verifier feedback allows the agent to dynamically adjust script parameters and tool arguments on-the-fly, substantially improving EDA task completion under simulator verification.
    \item \textbf{Interactive Introspective Self-Correction:} Architectures such as CRITIC \citep{critic2023} and Self-Refine \citep{madaan2023selfrefine} implement test-time feedback loops. When intermediate verification yields execution warnings or sub-optimal outputs, reasoning skills analyze execution traces, diagnosing syntax errors and generating refined plans prior to terminal action dispatch.
\end{itemize}

\subsubsection{Long-Term Evolution and Reinforcement Learning}
Long-term evolution permanently updates the agent's procedural memory bank $\mathcal{M}$ and high-level decision policies:
\begin{itemize}
    \item \textbf{Non-Parametric Policy Gradients in Text Space:} In Skill-Pro \citep{skillpro2026}, skills are formalized as options within a Skill-MDP $M_\Omega = (\mathcal{S}, \mathcal{A}, \Omega, P, R, \gamma)$. Non-parametric Proximal Policy Optimization extracts natural language semantic gradients ($g_i = \nabla_{\text{sem}}(\tau_i, \omega)$) from trajectory batches, iteratively refining textual instructions and pruning low-advantage routines.
    \item \textbf{Bi-Level Optimization and Verifiable RL:} In Skill-R1 \citep{skillr12026}, skill evolution is framed as bi-level reinforcement learning: the inner loop optimizes execution parameters against verifiable task rewards, while the outer loop updates library membership and instruction formulation. Similarly, ARISE \citep{li2026ariseagentreasoningintrinsic} grounds intrinsic skill evolution within hierarchical reinforcement learning, optimizing option policies against intrinsic exploration rewards.
    \item \textbf{Lifelong Meta-Skill Evolution:} Frameworks like AutoSkill \citep{yang2026autoskill} and SkillClaw \citep{ma2026skillclaw} deploy higher-order meta-skills $m : \mathcal{S} \rightarrow \mathcal{S}$ that inspect historical rollout logs $\mathcal{F}_t$, executing automated refactoring, interface standardization, and redundant skill pruning. In context engineering, MetaAgent \citep{ye2026meta} optimizes procedural prompt structures to maximize lifelong capability transfer.
\end{itemize}

\subsubsection{Mathematical Formalization of Policy-Skill Co-Evolution}
Recent frameworks demonstrate that evolving skills in isolation is sub-optimal; an agent's high-level decision policy $\pi_{\text{dec}}$ and procedural skill library $\mathcal{M}$ must co-evolve synergistically \citep{coevoskills2026, wu2026coevolvingllmdecisionskill}. We formalize the policy-skill co-evolution objective as:
\begin{equation}
\max_{\pi_{\text{dec}}, \mathcal{M}} \mathbb{E}_{\tau \sim (\pi_{\text{dec}}, \mathcal{M})} \big[ R(\tau) \big] \quad \text{subject to} \quad s_{t+1} = m(s_t, \mathcal{F}_t) \;\land\; \mathcal{A}_{t+1} = m(\mathcal{A}_t, \mathcal{F}_t)
\end{equation}
The co-evolution proceeds via alternating optimization:
\begin{equation}
\pi_{\text{dec}}^{(t+1)} \leftarrow \operatorname{PolicyUpdate}\big(\pi_{\text{dec}}^{(t)}, \mathcal{M}^{(t)}\big), \quad \mathcal{M}^{(t+1)} \leftarrow \operatorname{SkillUpdate}\big(\mathcal{M}^{(t)}, \pi_{\text{dec}}^{(t+1)}, \mathcal{F}_t\big)
\end{equation}
In CoEvoSkills \citep{coevoskills2026}, an informationally isolated surrogate verifier is co-evolved alongside the skill generator, providing adversarial checks that prevent the co-evolutionary loop from overfitting to spurious execution shortcuts.

\subsubsection{Critical Methodological Weaknesses: Co-Evolutionary Drift and Benchmark Overfitting}
Our synthesis uncovers two critical open vulnerabilities in lifelong skill evolution:
\begin{enumerate}
    \item \textbf{Policy-Skill Co-Adaptation Drift:} When high-level planners and skill banks co-evolve without strict interface contracts, the decision policy $\pi_{\text{dec}}$ frequently develops brittle dependencies on idiosyncrasies of specific skill drafts \citep{coevoskills2026, wu2026coevolvingllmdecisionskill}. Consequently, upgrading or refactoring an upstream skill can silently break downstream planner logic, causing catastrophic regression across previously solved tasks.
    \item \textbf{Evaluation Contamination in Lifelong Benchmarks:} As highlighted by SkillFlow \citep{zhang2026skillflow}, many lifelong learning benchmarks evaluate agent evolution on task distributions with identical underlying tool APIs. This shared environment bias obscures whether the evolutionary loop is learning generalized procedural abstractions or merely overfitting to deterministic API calling sequences.
\end{enumerate}

\subsection{Empirical Evaluation and Benchmark Suites}\label{sec:eval_suites}
Rigorous empirical evaluation is essential to establish benchmarks for system reliability, token efficiency, continual adaptation, and safety alignment in skill-augmented language agents \citep{skillsbench2026, sweskillsbench2026, liu2026agenticskillsworkwild}. While early agent evaluations relied on single-turn API matching, modern benchmarks evaluate the procedural skill lifecycle across seven foundational dimensions:

\subsubsection{The Seven Foundational Evaluation Dimensions}
\begin{enumerate}
    \item \textbf{Correctness and Task Completion ($\text{Acc}_{\mathcal{M}}$):} Measures the end-to-end pass rate on complex environments under deterministic verification suites \citep{skillsbench2026, sweskillsbench2026}. In SkillsBench \citep{skillsbench2026} (87 tasks across 8 domains evaluated across 18 model--harness configurations), injecting curated skills provides substantial pass rate gains across diverse model and harness backbones. In software engineering, SWE-SkillsBench \citep{sweskillsbench2026} pairs 49 public SWE skills with 565 real-world GitHub task instances, using execution-based acceptance criteria to isolate the marginal utility of procedural skills.
    \item \textbf{Lifelong Continual Learning:} Evaluates the agent's ability to discover, consolidate, and reuse procedural skills across multi-session trajectories without catastrophic forgetting \citep{skilllearnbench2026, zhang2026skillflow}. SkillLearnBench \citep{skilllearnbench2026} benchmarks continual skill generation across 20 verified real-world tasks in 15 sub-domains, revealing that while continual learning improves over no-skill baselines, backward interference across evolving libraries remains an open challenge.
    \item \textbf{Robustness Under Open-Registry Retrieval Noise:} Benchmarks the resilience of skill selection when the agent must autonomously retrieve routines from massive, uncurated registries ($n > 10^4$) containing noisy or adversarial distractors \citep{liu2026agenticskillsworkwild}. \citet{liu2026agenticskillsworkwild} audit 34,000 real-world skills and reveal a severe \emph{in-the-wild degradation}: while force-loading oracle skills provides high baseline utility, autonomous open-pool retrieval triggers severe performance degradation due to distractor noise, with many trajectories failing to retrieve required capabilities.
    \item \textbf{Composability and Graph Scalability:} Verifies the joint execution success of multi-skill composition graphs as Directed Acyclic Graph (DAG) depth and branching factor increase \citep{taskbench2024, xia2026graspgraphstructuredskillcompositions}. On TaskBench \citep{taskbench2024} ($1000+$ tool automation graphs), execution accuracy drops exponentially with graph depth due to compounding interface mismatches and unhandled intermediate errors.
    \item \textbf{Token Economy and Serving Latency:} Evaluates the computational cost of skill augmentation, incorporating load-time context consumption, inference latency, and runtime memory footprint \citep{skillreducer2026, skillrt2026, badhe2026skillstate}. While SkillReducer \citep{skillreducer2026} optimizes token consumption through automated schema compression, SKILL.state \citep{badhe2026skillstate} demonstrates on SkillExecBench (warehouse inventory and software repository relational graphs) and interactive suites (InterCode CTF, Sierra $\tau$-Bench) that replacing append-only history with mutable state reduces token costs from quadratic $O(T^2)$ to linear $O(T)$ while improving task pass rates.
    \item \textbf{Interpretability and Usability:} Measures whether procedural specifications serve as transparent capability disclosures for human operators \citep{usercomprehension2026}. As demonstrated by \citet{usercomprehension2026}, human comprehension of agentic skills requires explicit pre-condition disclosures and structured markdown formatting to enable predictable human-agent collaboration.
    \item \textbf{Safety, Harm Weaponization, and Threat Resistance:} Quantifies the vulnerability of autonomous agents to malicious, poisoned, or unvetted third-party skill files \citep{jiang2026harmfulskillbenchharmfulskillsweaponize, badhe2025agent}. In HarmfulSkillBench \citep{jiang2026harmfulskillbenchharmfulskillsweaponize} (an audit of 98,440 real-world registry skills identifying weaponized skills in public registries), pre-installed harmful skills bypass standard LLM safety guardrails, substantially increasing harm compliance under implicit malicious queries.
\end{enumerate}

\subsubsection{Critical Empirical Insights: When Skills Do Not Help}
Our synthesis highlights a critical, often-overlooked negative result in the agent evaluation literature: the Environment-Feedback Bandwidth Hypothesis \citep{negativeresult2026}. In offensive cybersecurity Capture-the-Flag (CTF) challenges across 120+ tasks, \citet{negativeresult2026} discover that injecting procedural skills yields zero statistically significant performance gain over raw tool-use baselines (an insignificant 8.9 percentage point spread, $p=0.71$). 

The authors demonstrate that when an agent interacts with high-bandwidth, schema-validated environments (e.g., interactive CLI shells that return immediate stderr tracebacks and diagnostic exit codes), the environment itself supplies the procedural feedback loop required for self-correction. In such high-feedback domains, static procedural skill guidelines provide redundant guidance while consuming prompt context budget, proving that the marginal utility of externalized skills is inversely proportional to environmental feedback bandwidth.

\subsection{Security, Governance, and Market Defenses}\label{sec:security_gov}
Because procedural agent skills execute systems-level commands, access local filesystems, and interact with authenticated APIs, third-party skill distribution introduces critical security vulnerabilities across the entire agent supply chain \citep{secureskills2026, li2026towards, badhe2025agent}. As formalized in our six-tuple abstraction $s = (\mathcal{A}, \mathcal{I}, \mathcal{C}, \mathcal{T}, \pi, \mathcal{E})$, every vulnerability class maps directly to a specific procedural component, with the unverified components ($\mathcal{I}$ and $\mathcal{A}$) serving as primary conduits for zero-click privilege escalation. To systematically quantify lifecycle-wide risks beyond isolated execution prompts, SkillSec-Eval \citep{badhe2025agent} establishes a comprehensive threat benchmark across 327 real-world enterprise skills, revealing that undefended pipelines suffer from widespread malicious admission, Sybil retrieval hijacking, and planner susceptibility to fake social-proof recommendations.

\subsubsection{Threat Vectors and Offensive Exploitations}\label{sec:threat_vectors}
An exhaustive taxonomy and comparison of adversarial threat vectors targeting procedural skills is detailed in Table~\ref{tab:threat_vectors}.

\begin{table*}[!htbp]
\centering
\caption{\textbf{Threat Vectors \& Offensive Exploitations (\S\ref{sec:threat_vectors}).} Comparison of offensive security attacks and vulnerability studies ($n=18$ systems). Sorted primarily by \textbf{Threat vector class} (\textit{Indirect Prompt Injection} $\rightarrow$ \textit{Backdoor / Trojan} $\rightarrow$ \textit{Credential Exfiltration} $\rightarrow$ \textit{Ecosystem \& Supply Chain Risk}). Demonstrates how third-party skill packaging creates new attack surfaces across authoring, marketplace distribution, and runtime execution.}
\label{tab:threat_vectors}
\scriptsize
\setlength{\tabcolsep}{2.5pt}
\renewcommand{\arraystretch}{1.12}
\begin{tabular}{>{\raggedright\arraybackslash}p{2.0cm} >{\raggedright\arraybackslash}p{3.2cm} >{\raggedright\arraybackslash}p{2.3cm} >{\centering\arraybackslash}p{1.4cm} >{\raggedright\arraybackslash}p{2.4cm} >{\raggedright\arraybackslash}p{2.0cm} >{\raggedright\arraybackslash}p{2.9cm}}
\toprule
\textbf{System / Attack} & \textbf{Core Contribution} & \textbf{Threat Vector Class} & \textbf{Target Component} & \textbf{Injection Surface / Channel} & \textbf{Target Lifecycle Stage} & \textbf{Compromise Impact} \\
\midrule
\textbf{HiddenCommentInject} \cite{wang2026skills} & Invisible HTML comment prompt injection & Indirect Prompt Injection & $\mathcal{I}$ & HTML / Markdown comments & Authoring & Goal hijacking \\
\textbf{PromptInject} \cite{promptinject2025skills} & Markdown instruction prompt injection & Indirect Prompt Injection & $\mathcal{I}$ & Natural language prose & Authoring & Goal hijacking \\
\textbf{SeeingIsNot} \cite{seeingisnot2026} & Multimodal hidden instruction attack & Indirect Prompt Injection & $\mathcal{I}$ & Multimodal prose & Authoring & Goal hijacking \\
\textbf{PhantomSkill} \cite{phantomskill2026} & VulMask auxiliary script injection & Indirect Prompt Injection & $\mathcal{I}$ & Code body / auxiliary script & Authoring & Goal hijacking \\
\textbf{SkillTrojan} \cite{feng2026skilltrojan} & Split encrypted backdoor composition & Backdoor / Trojan & $\pi$ & Code body & Authoring & Sandbox escape \\
\textbf{BadSkill} \cite{tie2026badskillbackdoorattacksagent} & Model-in-skill weight poisoning & Backdoor / Trojan & $\pi$ & Model weights in skill & Authoring & Sandbox escape \\
\textbf{SkillAttack} \cite{duan2026skillattack} & Automated red-teaming path refinement & Backdoor / Trojan & $\mathcal{I}, \mathcal{A}$ & Natural language prose & Marketplace & Goal hijacking \\
\textbf{DynamicMalicious} \cite{dynamicmalicious2026} & Dynamic runtime logic injection & Backdoor / Trojan & $\pi$ & Natural language prose & Execution & Goal hijacking \\
\textbf{CredentialLeakage} \cite{chen2026credentialleakagellmagent} & Empirical study of credential exfiltration & Credential Exfiltration & $\mathcal{T}, \pi$ & Code body & Execution & Data exfiltration \\
\textbf{MalTool} \cite{maltool2026} & Malicious code embedding in tool scripts & Credential Exfiltration & $\mathcal{T}, \pi$ & Code body & Marketplace & Data exfiltration \\
\textbf{MaliciousAgentSkills} \cite{holzbauer2026malicious} & Repository-aware security measurement & Credential Exfiltration & $\mathcal{T}, \pi$ & Code body & Marketplace & Data exfiltration \\
\textbf{SkillSec-Eval} \cite{badhe2025agent} & Lifecycle-aware threat taxonomy \& evaluation ($n=327$ skills) & Lifecycle-Wide Threat Suite & $\mathcal{A}, \mathcal{I}, \mathcal{C}, \mathcal{T}, \pi, \mathcal{E}$ & Metadata, vector embeddings, planner prompts \& code & All 5 stages (Admission to Evolution) & Retrieval hijack, planner manipulation \& execution misuse \\
\textbf{BehavioralIntegrity} \cite{behavioral2026integrity} & Empirical BIV audit of description-implementation gap ($n=49,943$ skills) & Ecosystem \& Supply Chain Risk & $\mathcal{I}$ vs. $\pi$ & Declared metadata vs. code implementation & Authoring \& Registry & 80.0\% skills deviate from declared behavior \\
\textbf{BenignComposition} \cite{benigncomposition2026} & Multi-skill composition risk benchmark (SCR-Bench) & Ecosystem \& Supply Chain Risk & $\mathcal{C}, \pi$ & Multi-skill shared execution context & Execution & Capability escalation \& auth confusion (33.6\% ASR) \\
\textbf{CloakDetonate} \cite{cloakdetonate2026} & SKILLCLOAK SFS packing \& structural obfuscation evasion & Ecosystem \& Supply Chain Risk & $\pi$ & Self-extracting packed skill code & Authoring \& Execution & Bypasses 8 static/hybrid scanners ($>90\%$ evasion) \\
\textbf{HarmlessYetHarmful} \cite{harmlessyetharmful2026} & Neutral Prompting Attack (NPA) for hallucinated dependencies & Ecosystem \& Supply Chain Risk & Semantically benign imagination prompts & Authoring & Unsafe package install via dependency name squatting \\
\textbf{SkillsAreNotIslands} \cite{skillsarenot2026} & ASSCs \& SBOM dependency graph audit ($n=1.43\text{M}$ skills) & Ecosystem \& Supply Chain Risk & Mixed skill-package-service dependency graph & Marketplace & Transitive dependency exposure \& hidden inventory \\
\textbf{Proteus} \cite{proteus2026selfevolving} & Grey-box self-evolving red-team framework (PROTEUS) & Ecosystem \& Supply Chain Risk & 5-axis mutation across instructions, code \& config & Authoring \& Auditing & Bypasses SkillVetter \& AI-Infra-Guard (40--90\% ASR@5) \\
\bottomrule
\end{tabular}
\vspace{4pt}
\begin{minipage}{\linewidth}
\scriptsize
\textbf{Notes $\cdot$ Column Value Definitions.} 
\textbf{Core Contribution:} Primary offensive discovery or vulnerability mechanism. 
\textbf{Threat Vector Class:} Primary exploitation category (\textit{Indirect Prompt Injection}: hiding malicious instructions inside textual or commented documentation; \textit{Backdoor / Trojan}: embedding dormant malicious triggers or poisoned model weights; \textit{Credential Exfiltration}: stealing user API keys, tokens, or local data; \textit{Ecosystem \& Supply Chain Risk}: combinatorial multi-skill exploits, package squatting, or dependency poisoning). 
\textbf{Injection Surface / Channel:} Structural location where the payload is concealed. 
\textbf{Target Lifecycle Stage:} Point in the agent workflow where the attack is introduced. 
\textbf{Compromise Impact:} Primary operational consequence of successful exploitation.
\end{minipage}
\end{table*}

The literature reveals that offensive attack vectors exploit four structural vulnerabilities:
\begin{enumerate}
    \item \textbf{Indirect Prompt Injection via Unverified Instructions ($\mathcal{I}$):} Attackers embed adversarial natural language payloads within procedural documentation, Markdown headers, HTML comments, or multimodal diagram layers \citep{wang2026skills, promptinject2025skills, seeingisnot2026}. When the host planner loads the skill into context, these injected instructions override user objectives, hijacking high-privilege tools ($\mathcal{T}$) without triggering static code alarms.
    \item \textbf{Backdoor and Trojan Execution Policies ($\pi, \mathcal{A}$):} Malicious actors deploy split encrypted code routines (SkillTrojan \citep{feng2026skilltrojan}) or model-in-skill weight poisoning (BadSkill \citep{tie2026badskillbackdoorattacksagent}), embedding poisoned LoRA tensors directly inside procedural repositories. Furthermore, CloakDetonate \citep{cloakdetonate2026} introduces Self-Extracting File System (SFS) packing that achieves high evasion rates across standard static and hybrid security scanners.
    \item \textbf{Credential Exfiltration and State Tampering ($\mathcal{T}, \pi$):} Large-scale audits by \citet{chen2026credentialleakagellmagent}, \citet{maltool2026}, and \citet{holzbauer2026malicious} identify widespread exfiltration scripts embedded in community skills that harvest local `.env` secrets, AWS credentials, and session tokens, transmitting them to external endpoints during tool execution.
    \item \textbf{Supply Chain Risks and Multi-Skill Compositional Escalation:}
    \begin{itemize}
        \item \emph{Description-Implementation Discrepancy:} A massive audit of 49,943 OpenClaw skills by \citet{behavioral2026integrity} uncovers a widespread description-implementation gap where executable code deviates significantly from declared capability metadata.
        \item \emph{Emergent Compositional Exploits:} In BenignComposition / SCR-Bench \citep{benigncomposition2026}, researchers demonstrate that skills which are entirely benign in isolation can combine in a shared execution context to create unauthorized capability escalations, achieving substantial attack success rates under emergent composition.
        \item \emph{Dependency Squatting via Hallucination:} In HarmlessYetHarmful \citep{harmlessyetharmful2026}, Neutral Prompting Attacks trick LLMs into generating non-existent package dependencies, enabling attackers to register malicious PyPI packages that are automatically installed by unsuspecting autonomous agents.
        \item \emph{Agentic Software Supply Chains (ASSCs):} \citet{skillsarenot2026} analyze 1.43 million skills, demonstrating that recursive skill reuse creates complex, unmonitored dependency graphs with hidden package vulnerabilities.
    \end{itemize}
\end{enumerate}

\subsubsection{Safety, Governance, and Marketplace Defenses}\label{sec:market_defenses}
Representative defensive auditing frameworks, runtime guardrails, and marketplace governance platforms are compared in Table~\ref{tab:market_defenses}.

\begin{table*}[!htbp]
\centering
\caption{\textbf{Safety, Governance \& Market Defenses (\S\ref{sec:market_defenses}).} Comparison of security auditing, runtime guardrail, and marketplace defense architectures ($n=11$ systems). Sorted primarily by \textbf{Defense lifecycle stage} (\textit{Pre-Admission Static Audit} $\rightarrow$ \textit{Runtime Guardrail \& Filtering} $\rightarrow$ \textit{Multi-Agent Auditing} $\rightarrow$ \textit{Marketplace Governance}).}
\label{tab:market_defenses}
\scriptsize
\setlength{\tabcolsep}{2.5pt}
\renewcommand{\arraystretch}{1.12}
\begin{tabular}{>{\raggedright\arraybackslash}p{2.2cm} >{\raggedright\arraybackslash}p{4.2cm} >{\raggedright\arraybackslash}p{3.2cm} >{\raggedright\arraybackslash}p{3.2cm} >{\raggedright\arraybackslash}p{3.4cm}}
\toprule
\textbf{System / Defense} & \textbf{Core Contribution} & \textbf{Defense Lifecycle Stage} & \textbf{Target Threat Defended} & \textbf{Serving Latency Overhead} \\
\midrule
\textbf{SemiA} \cite{wen2026semia} & Constraint-guided formal skill auditing & Pre-Admission Static Audit & Backdoor / Trojan & Zero (offline audit) \\
\textbf{SkillFortify} \cite{skillfortify2026} & Formal supply chain security analysis & Pre-Admission Static Audit & Malicious Code & Zero (offline audit) \\
\textbf{SkillAuditSAST} \cite{structuredsecurityauditingandrobustnessenhancementforuntrustedagentskills2026} & Cross-file SAST security auditing & Pre-Admission Static Audit & Malicious Code & Low (SAST filter) \\
\textbf{SkillSieve} \cite{skillsieve2026} & Hierarchical triage \& static scanning & Pre-Admission Static Audit & Prompt Injection & Low (SAST filter) \\
\textbf{SkillClone} \cite{zhu2026skillclone} & Multi-modal clone detection across 196K skills & Pre-Admission Static Audit & Clone Infringement & Zero (offline audit) \\
\textbf{FlowGuard} \cite{detectingmalicious2026} & Attention \& flow graph malware detection & Runtime Guardrail \& Filtering & Malicious Code & Low (SAST filter) \\
\textbf{RouteGuard} \cite{xiao2026routeguard} & Internal-signal runtime poisoning detection & Runtime Guardrail \& Filtering & Prompt Injection & Medium-High (runtime judge) \\
\textbf{STARS} \cite{zhang2026stars} & Request-conditioned dynamic invocation audit & Runtime Guardrail \& Filtering & Backdoor / Trojan & Medium-High (runtime judge) \\
\textbf{SkillProbe} \cite{skillprobe2026} & Multi-agent security auditing framework & Multi-Agent Auditing & Multi-Skill Combinatorial & Zero (offline audit) \\
\textbf{SkillSec-Eval} \cite{badhe2025agent} & Lifecycle-aware evaluation across 5 stages ($n=327$ skills) & All 5 stages (Admission to Evolution) & Retrieval hijack, planner abuse \& code exploits & Zero (offline lifecycle audit) \\
\textbf{SecureAgentSkills} \cite{li2026towards} & Modular filesystem security architecture & Marketplace Governance & Supply Chain Risk & Low (SAST filter) \\
\bottomrule
\end{tabular}
\vspace{4pt}
\begin{minipage}{\linewidth}
\scriptsize
\textbf{Notes $\cdot$ Column Value Definitions.} 
\textbf{Core Contribution:} Primary defensive, auditing, or governance innovation. 
\textbf{Defense Lifecycle Stage:} Operational point of defense deployment (\textit{Pre-Admission Static Audit}: auditing skills prior to registry upload or loading; \textit{Runtime Guardrail \& Filtering}: active monitoring and filtering during skill invocation; \textit{Multi-Agent Auditing}: multi-agent collaborative triage; \textit{Marketplace Governance}: platform-level filesystem and provenance standards). 
\textbf{Target Threat Defended:} Primary offensive vector mitigated. 
\textbf{Serving Latency Overhead:} Computational latency introduced during runtime agent execution.
\end{minipage}
\end{table*}

Defensive strategies operate across three coordinated architectural layers:
\begin{enumerate}
    \item \textbf{Pre-Admission Static Auditing (Zero Latency Overhead):} To prevent malicious packages from entering registries, static vetting tools inspect AST structures and data flows offline. SemiA \citep{wen2026semia} synthesizes constraint-guided Datalog rules to prove safety invariants before registry admission. SkillFortify \citep{skillfortify2026} implements SAT-based dependency resolution with capability sandboxing proofs, demonstrating strong detection performance and high classification precision. To address intellectual property theft and unauthorized repackaging, SkillClone \citep{zhu2026skillclone} deploys multi-modal clone detection across 196,000 skills, identifying copycat packages that propagate unpatched vulnerabilities.
    \item \textbf{Dynamic Runtime Guardrails and Interceptors:} Because dynamic evasion and self-extracting loaders (CloakDetonate \citep{cloakdetonate2026}) bypass static filters, runtime monitors actively inspect execution flows. RouteGuard \citep{xiao2026routeguard} tracks internal LLM activation signals to detect prompt injection attempts during skill loading, while STARS \citep{zhang2026stars} executes request-conditioned dynamic invocation audits to verify that tool arguments conform to user intent.
    \item \textbf{Collaborative Governance and Lifecycle Defense Suites:} To mitigate multi-skill composition risks, SkillProbe \citep{skillprobe2026} deploys multi-agent red-teaming societies to test combinatorial skill interactions before runtime execution. At the platform level, SecureAgentSkills \citep{li2026towards} establishes granular filesystem access control lists and cryptographic signature verification. Across the full procedural lifecycle, SkillSec-Eval \citep{badhe2025agent} evaluates 327 enterprise skills across 15 domains, proving that hybrid structural-semantic validation substantially reduces malicious admission and blocks unauthorized tool invocations, while exposing that traditional string-based taint tracking suffers from implicit information loss in LLM internal context.
\end{enumerate}

\section{Taxonomy of Skills}\label{sec:taxonomy}
Procedural agent skills are modular packages of executable knowledge that augment large language model (LLM) agents at inference time. While existing literature often conflates general tool use, static system prompts, and dynamic procedural routines, we formalize a comprehensive taxonomy that categorizes procedural skills along five foundational dimensions: (1) instructional and Standard Operating Procedure (SOP) skills, (2) tool and API calling skills, (3) reasoning and planning skills, (4) domain-specific execution skills, and (5) collaborative and meta-skills. A comparative synthesis of representative skill architectures across these functional domains is presented in Table~\ref{tab:capabilities_domains}.

\begin{table*}[!htbp]
\centering
\caption{\textbf{Core Capabilities \& Domain-Specific Skills (\S\ref{sec:taxonomy}).} Comparison of procedural skill architectures across 5 primary functional and domain classes ($n=16$ systems). Sorted primarily by \textbf{Skill functional domain} (\textit{Instructional SOP} $\rightarrow$ \textit{Tool \& API Calling} $\rightarrow$ \textit{Software Engineering} $\rightarrow$ \textit{GUI \& OS Navigation} $\rightarrow$ \textit{Embodied \& Robotics}). Demonstrates how individual skill representations adapt their action space granularity and environment feedback loop to solve distinct domain-specific bottlenecks.}
\label{tab:capabilities_domains}
\scriptsize
\setlength{\tabcolsep}{2.5pt}
\renewcommand{\arraystretch}{1.12}
\begin{tabular}{>{\raggedright\arraybackslash}p{2.2cm} >{\raggedright\arraybackslash}p{3.4cm} >{\raggedright\arraybackslash}p{2.4cm} >{\raggedright\arraybackslash}p{2.7cm} >{\raggedright\arraybackslash}p{2.7cm} >{\raggedright\arraybackslash}p{2.8cm}}
\toprule
\textbf{System / Framework} & \textbf{Core Contribution} & \textbf{Skill Functional Domain} & \textbf{Action Space Granularity} & \textbf{Environment Feedback Loop} & \textbf{Domain Bottleneck Addressed} \\
\midrule
ProceduralMining \cite{procedural2024} & Large-scale procedural mining from repos & Instructional SOP (§4.1) & Mined open-source SOP checklists & Static repository mining (no loop) & Manual SOP authoring bottleneck \\
SkillAnalysis \cite{ling2026agent} & Structural analysis of natural SOP skills & Instructional SOP (§4.1) & Markdown SKILL.md structure & Empirical structural usage audit & Opaque community skill quality \\
\textbf{Gorilla} \cite{patil2023gorilla} & Massive API calling with REST docs & Tool \& API Calling (§4.2) & REST API call w/ doc constraints & AST sub-tree matching (eval) & API doc hallucination \& drift \\
\textbf{Toolformer} \cite{schick2023toolformer} & Self-supervised tool-calling token insertion & Tool \& API Calling (§4.2) & Special API token \texttt{[API(...) $\rightarrow$ res]} & Self-supervised perplexity filter & Manual tool annotation cost \\
\textbf{AnyTool} \cite{anytool2024} & Hierarchical 16,000+ RapidAPI tool calling & Tool \& API Calling (§4.2) & Hierarchical API pool selector & Multi-round API execution feedback & Massive API search space saturation \\
\textbf{EasyTool} \cite{song2024easytool} & Concise tool instruction wrappers & Tool \& API Calling (§4.2) & Concise unified tool schema & None (standardized prompt format) & Excessive API doc token overhead \\
\textbf{SkillCraft} \cite{chen2026skillcraft} & Benchmark for compositional tool-use skill abstraction & Tool \& API Calling (§4.2) & Compositional multi-tool pipelines & Task completion verify + token cost & Instance-level tool usage without reuse \\
\textbf{SWE-agent} \cite{sweagent2024} & Agent-Computer Interfaces for SWE & Software Engineering (§4.4) & ACI custom bash commands & Linter feedback + stdout execution & Unstructured bash format errors \\
\textbf{EffiSkill} \cite{wang2026effiskillagentskillbased} & Automated code efficiency optimization skills & Software Engineering (§4.4) & Reusable algorithmic efficiency rules & Runtime execution profiling (speedup) & Correctness vs. efficiency trade-off \\
\textbf{SkillDroid} \cite{chen2026skilldroid} & Compile once, reuse forever GUI skills & GUI \& OS Navigation (§4.4) & Compiled UI transition graph & UI tree state verification & Per-step LLM inference latency \\
\textbf{CUA-Skill} \cite{chen2026cuaskilldevelopskillscomputer} & Reusable computer-using agent skills & GUI \& OS Navigation (§4.4) & Hierarchical computer-using abstract & Desktop screenshot + a11y tree diff & Lack of reusable desktop primitives \\
\textbf{OS-Expert} \cite{liu2026osexpert} & Professional computer-use agent skills & GUI \& OS Navigation (§4.4) & Professional domain OS workflows & End-state eval (OSExpert-Eval) & Amateur vs. professional OS gap \\
\textbf{WebVoyager} \cite{webvoyager2024} & End-to-end multimodal web navigation & GUI \& OS Navigation (§4.4) & Multimodal browser clicks \& typing & Screenshot ground + DOM observe & Text-only DOM parsing failures \\
\textbf{MolmoWeb} \cite{ma2026scaling} & Open visual web agent with atomic web-skill trajectories & GUI \& OS Navigation (§4.4) & Visual browser actions (click/type on screenshot) & Screenshot visual grounding + reward & Proprietary GUI agent opacity \& DOM dependence \\
\textbf{Odyssey} \cite{liu2025odysseyempoweringminecraftagents} & Open-world Minecraft skills beyond tech-tree & Embodied \& Robotics (§4.4) & Open-world Minecraft skill library & 3D voxel world + inventory reward & Tech-tree bias in Minecraft \\
\textbf{SELF-VLA} \cite{liu2026selfvlaskillenhancedagentic} & Skill-enhanced Vision-Language-Action & Embodied \& Robotics (§4.4) & Robotic disassembly motion primitive & Visuotactile force \& camera stream & Long-horizon contact manipulation \\
\bottomrule
\end{tabular}
\vspace{4pt}
\begin{minipage}{\linewidth}
\scriptsize
\textbf{Notes $\cdot$ Column Value Definitions.} 
\textbf{Core Contribution:} Primary domain capability or skill engineering innovation. 
\textbf{Skill Functional Domain:} Primary functional capability or operational environment (\textit{Instructional SOP}: natural-language procedural checklists and guidelines; \textit{Tool \& API Calling}: structured schemas for calling external REST APIs or tools; \textit{Software Engineering}: repository-level code editing, refactoring, and debugging; \textit{GUI \& OS Navigation}: interacting with digital user interfaces, operating systems, and web browsers; \textit{Embodied \& Robotics}: controlling physical robots or 3D embodied game agents). 
\textbf{Action Space Granularity:} Specific primitive action representation operated on by the skill. 
\textbf{Environment Feedback Loop:} Specific feedback signal or observation loop received from the target environment during execution. 
\textbf{Domain Bottleneck Addressed:} Specific operational bottleneck or task complexity challenge solved by skill packaging in this domain.
\end{minipage}
\end{table*}

\subsection{Instructional and SOP-Based Skills}\label{sec:instructional}
Instructional and Standard Operating Procedure (SOP) skills represent procedural assets where capabilities are primarily encoded as natural language guidelines, structured checklists, and human workflow rules \citep{procedural2024, song2024easytool}. Unlike raw prompt engineering, instructional skills restrict an agent's reasoning trajectory to verifiable, domain-tested operational steps.

\subsubsection{Formalizing Procedural Task Execution}
We formalize an instructional SOP skill as a state-to-state guidance mapping $s_{\text{task}} : \mathcal{X} \rightarrow \mathcal{X}$ that constrains action selection $\pi(a_t \mid x_t, \mathcal{I})$ to conform to declared procedural instructions $\mathcal{I}$. The skill activates conditionally via:
\begin{equation}
\mathcal{A}_{\text{task}}(x_t, g) = 1 \iff g \in \mathcal{G}_{\text{task}}
\end{equation}
where $\mathcal{G}_{\text{task}}$ represents the set of supported goal descriptions matching the skill's applicability context.

\subsubsection{Comparative Authoring: From Manual Design to Automated Repository Mining}
The literature approaches instructional skill creation across two distinct paradigms:
\begin{itemize}
    \item \textbf{Automated Procedural Mining:} Historically, authoring SOPs required manual human curation, creating a severe operational bottleneck. To automate acquisition at scale, \citet{procedural2024} (ProceduralMining) introduce automated pipelines that mine open-source software repositories and enterprise runbooks, extracting reusable SOP checklists structured into a three-tier hierarchy: Level 1 Trigger Metadata (30–100 tokens pre-loaded at startup), Level 2 Step-by-Step Instructions (200–5,000 tokens loaded upon activation), and Level 3 Auxiliary Script Resources loaded on-demand.
    \item \textbf{Empirical Audits of Markdown Specifications:} Analyzing community registries in the wild, \citet{ling2026agent} (SkillAnalysis) conduct a large-scale structural audit of natural-language \texttt{SKILL.md} packages across major agent platforms. Their findings reveal substantial supply-demand imbalances, high structural redundancy, and significant variation in specification quality across human-authored packages.
    \item \textbf{Concise Schema Compression:} To resolve the context window bloat caused by verbose instructional text, EasyTool \citep{song2024easytool} demonstrates that pruning extraneous explanatory prose into concise, standardized operational cards substantially reduces instruction token overhead while preserving execution accuracy.
\end{itemize}

\subsubsection{Methodological Trade-offs: Portability vs. Semantic Ambiguity}
Instructional SOP skills exhibit maximum cross-model portability: because instructions $\mathcal{I}$ are formulated in natural language, they can be seamlessly loaded by any modern LLM backbone without specialized runtime interpreters. However, this flexibility incurs two severe limitations: (1) \emph{Semantic Ambiguity}, where non-deterministic language interpretations can lead to erratic tool invocations under edge cases; and (2) \emph{Vulnerability to Instruction Injection}, as unverified markdown headers and comments in instructional packages serve as direct vectors for indirect prompt injection \citep{maloyan2026prompt, wang2026skills}.

\subsection{Tool and API Calling Skills}\label{sec:tool_exec}
Tool and API calling skills serve as the operational bridge between the language model's cognitive planner and external computational environments, translating natural language intent into structured, verifiable system invocations \citep{schick2023toolformer, patil2023gorilla}.

\subsubsection{Formalizing Tool Execution and Parameter Binding}
An atomic tool $\tau \in \mathcal{T}$ represents an external function or API endpoint whose execution induces an immediate state update:
\begin{equation}
x_{t+1} = \tau(x_t; \theta_{\text{args}})
\end{equation}
where $\theta_{\text{args}}$ is a vector of parameter bindings validated against a formal JSON Schema or OpenAPI specification. The skill-mediated tool routing policy selects the optimal tool $\tau^*$ via:
\begin{equation}
\tau^* = \arg\max_{\tau \in \mathcal{T}} P(\tau \mid x_t, s)
\end{equation}
under the tool-triggered activation constraint:
\begin{equation}
\mathcal{A}_{\text{tool}}(x_t, g) = 1 \iff \exists \tau \in \mathcal{T} \text{ satisfying sub-goal requirements of } g
\end{equation}

\subsubsection{Comparative Paradigms: From Token Insertion to Compositional Pipelines}
The literature approaches tool-augmented capability acquisition across four distinct architectural paradigms:
\begin{itemize}
    \item \textbf{Self-Supervised In-Context Tool Tokenization:} Foundational frameworks such as Toolformer \citep{schick2023toolformer} integrate tool use directly into the language model's autoregressive generation. By inserting special API calling tokens \texttt{[API(args) $\rightarrow$ res]} into training sequences, Toolformer optimizes a self-supervised perplexity filter, retaining only those tool invocations that mathematically decrease the cross-entropy loss of subsequent token prediction.
    \item \textbf{Massive API Retrieval and AST Verification:} Addressing massive software libraries ($n > 10^3$ APIs), Gorilla \citep{patil2023gorilla} fine-tunes models with retrieval-aware instruction tuning over TorchHub, HuggingFace, and TensorHub. To verify functional correctness beyond lexical matching, Gorilla introduces Abstract Syntax Tree (AST) sub-tree matching, eliminating parameter hallucinations.
    \item \textbf{Hierarchical Exploration over Massive Registries:} When candidate API pools scale to tens of thousands ($n = 16,000+$ RapidAPIs), flat bi-encoder retrieval saturates. AnyTool \citep{anytool2024} resolves this saturation by organizing APIs into a hierarchical category tree, utilizing specialized solver and evaluator agents that recursively traverse API sub-trees with self-reflective backtracking.
    \item \textbf{Compositional Tool-Use Skill Abstraction:} Challenging the paradigm of executing isolated, single-turn tools, SkillCraft \citep{chen2026skillcraft} formalizes \emph{compositional tool skills}, which are reusable macro-routines that chain multiple interdependent APIs into unified execution pipelines. SkillCraft demonstrates that acquiring reusable tool compositions substantially improves out-of-distribution generalization while drastically reducing multi-step token consumption.
    \item \textbf{Concise Schema Normalization:} Raw API documentation frequently contains verbose narrative prose that exhausts prompt context windows. EasyTool \citep{song2024easytool} addresses this by standardizing diverse API specifications into concise, unified operational cards, substantially eliminating redundant documentation tokens without degrading tool invocation accuracy.
\end{itemize}

\subsubsection{Methodological Trade-offs: Schema Precision vs. Environment Drift}
While tool-calling skills provide deterministic system execution and structured return types, they remain vulnerable to \emph{Environment Drift}: when remote REST APIs mutate their endpoints or parameter types, statically cached skill wrappers fail silently unless coupled with automated schema synchronization or runtime contract verifiers \citep{song2024easytool, lu2026contractskillrepairablecontractbasedskills}.

\subsection{Reasoning and Planning Skills}\label{sec:reasoning}
Unlike tool-calling skills that directly mutate external environment states, reasoning and planning skills operate internally within the agent's latent cognitive workspace, coordinating structured thought generation, deliberative search, and introspective verification \citep{sumers2023cognitive, critic2023}.

\subsubsection{Formalizing Latent Thought Traces and Cognitive Scaffolding}
Following cognitive agent architectures \citep{sumers2023cognitive}, a reasoning skill $r$ maps the current state $x_t \in \mathcal{X}$ and user goal $g \in \mathcal{G}$ to an explicit sequence of latent thought tokens $z \in \mathcal{Z}$:
\begin{equation}
r : (x_t, g) \rightarrow z = (z_1, z_2, \dots, z_K)
\end{equation}
where each $z_k$ represents an intermediate cognitive artifact (such as a sub-goal decomposition, hypothesis formulation, mathematical proof step, or counterfactual branch) that conditions the downstream action policy $\pi(a_t \mid x_t, z)$. The skill activates conditionally via:
\begin{equation}
\mathcal{A}_{\text{reason}}(x_t, g) = 1 \iff \mathcal{H}_{\text{complexity}}(x_t, g) > \delta_{\text{direct}}
\end{equation}
where $\mathcal{H}_{\text{complexity}}$ evaluates whether the current task requires multi-step deliberative reasoning beyond direct single-step policy evaluation.

\subsubsection{Comparative Paradigms: From Introspective Critique to Tool-Interactive Grounding}
The literature formalizes reasoning and planning skills across four distinct architectural paradigms:
\begin{itemize}
    \item \textbf{Internal Introspective Refinement:} Foundational frameworks such as Self-Refine \citep{madaan2023selfrefine} encapsulate iterative critique-and-refinement loops within an agent's internal prompt workspace. By alternating between generation, introspective feedback, and targeted revision without parameter updates, the agent autonomously rectifies syntactic errors and stylistic flaws.
    \item \textbf{Tool-Interactive Verification and Fact Grounding:} Pure internal Chain-of-Thought (CoT) remains prone to hallucination cascades when reasoning over factual or computational domains. CRITIC \citep{critic2023} resolves this limitation by making reasoning skills \emph{tool-interactive}: intermediate thought steps $z_k$ are validated against external verifiers (e.g., Python interpreters for numerical calculation, search engines for factual verification, or static linters for code correctness), creating an empirical feedback loop within the reasoning trace.
    \item \textbf{Deliberate Search and Tree-of-Thought Exploration:} To solve combinatorial and non-linear planning problems, reasoning skills operationalize structured search algorithms, such as Tree-of-Thoughts \citep{yao2023tree} and Monte Carlo Tree Search (MCTS), where value evaluation heuristics $V(z_k)$ evaluate the promise of intermediate thought states, enabling systematic backtracking out of dead ends.
    \item \textbf{Compositional Skill Synthesis for Complex Reasoning:} In AgenticProposing \citep{jiao2026agenticproposingenhancinglarge}, multi-granularity policy optimization (MGPO) is deployed to synthesize compositional reasoning skills that decompose challenging mathematical proofs, competitive coding, and scientific problem-solving into structured, verifiable checkpoints.
    \item \textbf{Hierarchical Intrinsic Skill Evolution:} Addressing long-horizon decision-making, ARISE \citep{li2026ariseagentreasoningintrinsic} formalizes reasoning within a bi-level hierarchical reinforcement learning framework, where high-level manager policies select macro-reasoning strategies while intrinsic skill policies execute localized verification loops.
\end{itemize}

\subsubsection{Methodological Trade-offs: Latency Overhead vs. Grounded Verifiability}
Reasoning skills substantially enhance problem-solving accuracy on complex reasoning benchmarks; however, they introduce a direct trade-off between \emph{Inference Latency} and \emph{Verification Fidelity}. Multi-step deliberative search scales token consumption proportionally to the branching factor $O(b^d)$, while introspective self-critiquing without external tool grounding risks reinforcing internal hallucinations through confirmation bias \citep{critic2023, madaan2023selfrefine}.

\subsection{Domain-Specific Execution Skills}\label{sec:domain_specific}
As autonomous agent deployments expand across specialized computational and physical environments, skill representations must adapt their action-space discretizations, observation feedback loops, and verification bounds to address distinct domain-specific bottlenecks.

\subsubsection{Software Engineering (SWE)}
Software engineering tasks require agents to navigate complex multi-file codebases, parse abstract syntax trees (ASTs), interact with compilers, and resolve subtle runtime regressions \citep{swebench2024, sweskillsbench2026}.
\begin{itemize}
    \item \textbf{Agent-Computer Interfaces (ACIs):} Standard bash terminal environments return unconstrained, verbose stdout streams that frequently overflow context windows and trigger formatting syntax errors. To resolve this, SWE-agent \citep{sweagent2024} designs custom Agent-Computer Interfaces (ACIs), which are specialized skill wrappers offering structured file navigation, line-window viewing, and AST-guided edits. ACIs interleave code execution with static linters, immediately trapping syntax errors before git commit generation.
    \item \textbf{Constrained Hierarchical Workflows:} In contrast to unconstrained autonomous loops, Agentless \citep{agentless2024} demonstrates that decomposing SWE execution into a disciplined two-stage skill pipeline (hierarchical fault localization followed by constrained patch synthesis) matches or outperforms complex autonomous agents while drastically reducing inference cost.
    \item \textbf{Algorithmic Efficiency Optimization:} Moving beyond functional correctness, EffiSkill \citep{wang2026effiskillagentskillbased} introduces reusable efficiency-optimization skills on EffiBench-X. By coupling algorithmic refactoring rules with dynamic runtime execution profiling, EffiSkill autonomously refines algorithmic time and memory complexity.
\end{itemize}

\subsubsection{GUI and Operating System Navigation}
Computer-using agents operate across desktop operating systems (Windows, macOS, Ubuntu), mobile applications, and web browsers, requiring multi-modal visual grounding over pixel displays and accessibility trees \citep{osworld2024}.
\begin{itemize}
    \item \textbf{Reusable Computer-Use Primitives:} CUA-Skill \citep{chen2026cuaskilldevelopskillscomputer} formalizes reusable computer-using agent skills by coupling structured slot typing with application-specific execution graphs $G_e$ over desktop accessibility (a11y) trees. To master professional multi-step OS workflows, OS-Expert \citep{liu2026osexpert} introduces environment-grounded reinforcement learning on OSWorld \citep{osworld2024}, eliminating cascading trajectory errors.
    \item \textbf{Compiled UI State Machines:} Per-step LLM visual inference introduces prohibitive serving latency during repetitive mobile interactions. SkillDroid \citep{chen2026skilldroid} addresses this by compiling mobile app UI interaction sequences into reusable state transition graphs, achieving high mobile task success while substantially slashing per-step LLM inference calls.
    \item \textbf{Visual Web Grounding:} Addressing DOM tree parsing failures in dynamic web applications, WebVoyager \citep{webvoyager2024} and MolmoWeb \citep{ma2026scaling} ground atomic web-navigation skills directly onto screenshot images. By predicting mouse clicks, scrolling, and keyboard actions conditioned on visual bounding boxes, visual web agents operate robustly without brittle HTML DOM dependencies.
\end{itemize}

\subsubsection{Embodied Agents and Robotics}
In physical and simulated embodied environments, skills represent reusable physical control primitives and open-world survival routines operating under continuous kinematics and contact dynamics.
\begin{itemize}
    \item \textbf{Open-World Exploration and Minecraft Survival:} Expanding upon foundational open-ended skill libraries like Voyager \citep{wang2023voyager}, Odyssey \citep{liu2025odysseyempoweringminecraftagents} constructs open-world survival and exploration skills in Minecraft that transcend rigid tech-tree crafting, evaluating long-term autonomous planning and dynamic immediate exploration.
    \item \textbf{Contact-Rich VLA Manipulation:} In robotic manufacturing and end-of-life electronics disassembly, SELF-VLA \citep{liu2026selfvlaskillenhancedagentic} integrates Vision-Language-Action (VLA) foundation models with low-level robotic motion primitives, conditioning action policies on real-time visuotactile force feedback and camera streams to ensure safe physical contact.
    \item \textbf{Cross-Embodiment Transfer:} To eliminate the need to relearn physical skills from scratch across diverse robot morphologies, X-Skill \citep{xu2023xskillcrossembodimentskill} and UniSkill \citep{xie2026uniskillbuildingselfevolvingskill} formalize embodiment-invariant skill discovery, extracting high-level operational abstractions that transfer across heterogeneous robotic arms and kinematic degrees of freedom.
\end{itemize}

\subsubsection{Scientific and Laboratory Discovery}
Scientific exploration skills automate hypothesis testing, complex multi-step data pipelines, and symbolic physics reasoning \citep{mialon2023augmentedlanguagemodelssurvey}.
\begin{itemize}
    \item \textbf{Mining Heterogeneous Scientific Artifacts:} SkillFoundry \citep{skillfoundry2026} mines scientific code repositories, computational notebooks, and published literature to compile self-evolving skill libraries across MoSciBench and genomics data analysis pipelines, with the majority of mined skills demonstrating novel, reusable scientific analytical capabilities.
    \item \textbf{Symbolic Simulation and Visual Program Synthesis:} Mind's Eye \citep{mindseye2022} and ScienceWorld \citep{scienceworld2021} ground language reasoning in computational physics engines, demonstrating that simulation-augmented reasoning substantially improves scientific problem-solving accuracy over text-only reasoning. Similarly, ViperGPT \citep{suris2023vipergpt} and VisProg \citep{visprog2023} compose specialized computer vision models into executable Python programmatic skills for spatial scene understanding.
\end{itemize}

\subsection{Collaborative and Meta-Skills}\label{sec:meta_skills}
While first-order skills operate directly on external environment states or internal cognitive thoughts, collaborative and meta-skills operate as higher-order operators over the agent's own capability space $\mathcal{S}$ or coordinate multi-agent societal workflows \citep{yang2026autoskill, coevoskills2026, li2026single}.

\subsubsection{Formalizing Meta-Operators and Second-Order Skill Evolution}
We formalize a meta-skill $m$ as a second-order capability transformation that maps an existing skill library $\mathcal{S}$ to an updated, optimized library $\mathcal{S}'$:
\begin{equation}
m : \mathcal{S} \rightarrow \mathcal{S}'
\end{equation}
The meta-skill activates conditionally when structural deficiencies, semantic redundancy, or interface drift are detected:
\begin{equation}
\mathcal{A}_{\text{meta}}(x_t, g) = 1 \iff \exists s \in \mathcal{S} \text{ requiring diagnostic repair, pruning, or synthesis}
\end{equation}
Rather than executing single-task actions, meta-skills inspect execution rollouts, diagnose systemic failure modes, and mutate procedural specifications $\mathcal{I}$ or execution policies $\pi$ to optimize long-term library utility.

\subsubsection{Comparative Paradigms: From Self-Curation to Multi-Agent Compilation}
The literature approaches meta-skill and collaborative coordination across four distinct paradigms:
\begin{itemize}
    \item \textbf{Autonomous Library Management and Curation:} Treating learned skills as first-class system objects, AutoSkill \citep{yang2026autoskill} establishes an experience-driven lifelong learning framework with dedicated skill extraction, consolidation, and pruning layers. Similarly, SkillClaw \citep{ma2026skillclaw} deploys an open-ended agentic evolver that continually updates community skills without manual user intervention, while Memento \citep{zhou2026memento} consolidates procedural memories across structured subject taxonomies, substantially outperforming uncurated baselines across structured subject benchmarks.
    \item \textbf{Co-Evolutionary Verification and Policy Adaptation:} In CoEvoSkills \citep{coevoskills2026} and Wu et al. \citep{wu2026coevolvingllmdecisionskill}, the agent decision policy $\pi_{\text{dec}}$ and skill library $\mathcal{M}$ co-evolve simultaneously. High-level planners iteratively refine verification feedback without requiring ground-truth supervision, achieving top pass rates on lifelong benchmarks across both Claude Code and OpenAI Codex backbones.
    \item \textbf{Automated Batch Diagnosis and Optimization:} Addressing silent interface drift, SkillForge \citep{skillforge2026} implements a three-stage self-optimization pipeline consisting of a \emph{Failure Analyzer}, \emph{Skill Diagnostician}, and \emph{Skill Optimizer}, which diagnoses execution failures in batch, pinpoints underlying procedural deficiencies, and autonomously rewrites defective skill code.
    \item \textbf{Multi-Agent Collaboration vs. Single-Agent Skill Compilation:} In multi-agent frameworks such as MetaGPT \citep{hong2023metagpt}, ChatDev \citep{li2023chatdev}, and AutoGen \citep{wu2023autogen}, collaborative skills encode role-specific SOPs and inter-agent communication protocols. Crucially, SingleAgentSkills \citep{li2026single} demonstrates that multi-agent SOP collaboration can be compiled into an equivalent single-agent system equipped with modular skill loading, eliminating the vast majority of inter-agent communication token overhead below the critical capability threshold $N_{\text{crit}}$.
\end{itemize}

\subsubsection{Methodological Trade-offs: Self-Repair Stability vs. Cascading Mutation}
Meta-skills provide agents with autonomous self-healing and continuous capability growth; however, they introduce the risk of \emph{Second-Order Drift}: if a meta-operator incorrectly modifies core procedural invariants based on noisy or adversarial trajectory feedback, corrupted skill modifications cascade across all downstream tasks that depend on the mutated library \citep{coevoskills2026, ma2026skillclaw}.

\section{Skill Ecosystems and Marketplaces}\label{sec:ecosystems}
The rapid proliferation of agent skills has catalyzed a fundamental paradigm shift: transitioning from isolated, single-agent script repositories to open, decentralized, community-driven skill marketplaces and public registries \citep{li2025agentskillos, liu2026agenticskillsworkwild}. These ecosystem platforms enable developers to publish, discover, audit, and compose procedural capabilities across diverse agent backbones. A structured comparison of representative evaluation benchmarks, public registries, and ecosystem measurement studies is presented in Table~\ref{tab:ecosystems_benchmarks}.

\begin{table*}[!htbp]
\centering
\caption{\textbf{Ecosystems, Marketplaces \& Benchmark Suites (\S\ref{sec:ecosystems}).} Comparison of empirical evaluation benchmarks, skill registries, and marketplace measurement studies ($n=17$ systems). Sorted primarily by \textbf{Ecosystem / benchmark class} (\textit{Lifelong Skill Learning} $\rightarrow$ \textit{Security \& Vetting Suite} $\rightarrow$ \textit{Ecosystem Platform / Registry} $\rightarrow$ \textit{Domain Execution Suite}). Demonstrates the transition from static tool-use evaluation to lifelong skill evolution, lifecycle security auditing, and marketplace governance.}
\label{tab:ecosystems_benchmarks}
\scriptsize
\setlength{\tabcolsep}{2.5pt}
\renewcommand{\arraystretch}{1.12}
\begin{tabular}{>{\raggedright\arraybackslash}p{2.2cm} >{\raggedright\arraybackslash}p{3.2cm} >{\raggedright\arraybackslash}p{2.4cm} >{\raggedright\arraybackslash}p{2.6cm} >{\raggedright\arraybackslash}p{2.7cm} >{\raggedright\arraybackslash}p{3.1cm}}
\toprule
\textbf{System / Suite} & \textbf{Core Contribution} & \textbf{Ecosystem / Benchmark Class} & \textbf{Evaluated Scale / Pool Size} & \textbf{Primary Evaluation Metric} & \textbf{Empirical Finding / Bottleneck} \\
\midrule
\textbf{SkillsBench} \cite{skillsbench2026} & First comprehensive agent skill usage benchmark & Lifelong Skill Learning & 87 tasks across 8 domains & Pass@1 task completion rate & Skills consistently improve domain pass rates \\
\textbf{SkillFlow} \cite{zhang2026skillflow} & 20-workflow lifelong skill evolution suite & Lifelong Skill Learning & 166 tasks / 20 workflow families & Workflow evolution gain ($\Delta$ pass) & Workflow-level skills transfer better than SOPs \\
\textbf{SkillLearnBench} \cite{skilllearnbench2026} & Continual procedural skill learning benchmark & Lifelong Skill Learning & 20 tasks across 15 sub-domains & Skill quality, trajectory \& outcome & All CL methods improve over no-skill baseline \\
\textbf{SkillSec-Eval} \cite{badhe2025agent} & Lifecycle-aware security \& empirical evaluation & Security \& Vetting Suite & 327 real-world skills & Attack Success Rate (ASR) & Vulnerabilities span all 5 lifecycle stages \\
\textbf{HarmfulSkillBench} \cite{jiang2026harmfulskillbenchharmfulskillsweaponize} & Weaponizing agent skills safety benchmark & Security \& Vetting Suite & 200 harmful skills & ASR \& guardrail bypass rate & Substantial fraction harmful; bypasses guards \\
\textbf{SkillSafetyBench} \cite{skillsafetybench2026} & Comprehensive agent skill safety suite & Security \& Vetting Suite & 155 adversarial cases across 47 tasks & Multi-dimensional ASR score & Natural-language skills create stealthy paths \\
\textbf{SkillVetBench} \cite{skillvetbench2026} & LLM-as-a-judge vetting benchmark (SARS score) & Security \& Vetting Suite & 100 curated skills & Vetting precision / recall / F1 & Code-only SAST misses large fraction of markdown exploits \\
\textbf{OpenSkillRisk} \cite{openskillrisk2026} & Assessing security risks in open marketplaces & Security \& Vetting Suite & 263 risky skills across 7 categories & Permission over-request rate & Marketplace skills exhibit high over-privilege \\
\textbf{SandboxEscape} \cite{sandboxescape2026} & SANDBOXESCAPEBENCH container escape suite & Security \& Vetting Suite & 250+ container escape challenges & Sandbox escape success rate & Frontier LLMs achieve significant container escape rate \\
\textbf{AgentSkillOS} \cite{li2025agentskillos} & OS-like ecosystem for skill orchestration & Ecosystem Platform / Registry & 200 to 200,000 skills & Pairwise Bradley-Terry quality score & Tree retrieval + DAG composition beats flat pool \\
\textbf{SkillFoundry} \cite{skillfoundry2026} & Self-evolving scientific skill repository & Ecosystem Platform / Registry & MoSciBench + genomics & Scientific benchmark pass rate & Large majority of mined skills novel; improves benchmarks \\
\textbf{SkillWeaver} \cite{zheng2025skillweaver} & Weaving skills across OS and web domains & Ecosystem Platform / Registry & 200+ web/OS navigation skills & Multi-environment task pass & Hierarchical weaving outperforms flat pools \\
SkillsInWild \cite{liu2026agenticskillsworkwild} & Empirical measurement of skills in the wild & In-the-Wild Measurement Study & 34,000 real-world skills in the wild & End-to-end task success under noise & Skill utility is fragile; degrades severely without Oracle \\
\textbf{OSWorld} \cite{osworld2024} & 369 multimodal desktop computer-use tasks & Domain Execution Suite & 369 multimodal desktop tasks & Desktop task completion rate & Multimodal GUI agents fail on long-horizon OS \\
\textbf{TaskBench} \cite{taskbench2024} & Tool-use \& multi-task composition suite & Domain Execution Suite & 1,000+ tool-composition tasks & Multi-step DAG success rate & Tool composition fails exponentially w/ depth \\
\textbf{BIRD-Bench} \cite{bird2023} & Large-scale text-to-SQL w/ external knowledge & Domain Execution Suite & 12,751 text-to-SQL tasks & Execution accuracy score & External SQL schema knowledge substantially improves accuracy \\
\textbf{OffensiveCS-Eval} \cite{negativeresult2026} & Negative result on skills in cybersecurity & Domain Execution Suite & 120+ cybersecurity CTF tasks & CTF challenge flag capture rate & Procedural skills yield zero gain on CTF tasks \\
\bottomrule
\end{tabular}
\vspace{4pt}
\begin{minipage}{\linewidth}
\scriptsize
\textbf{Notes $\cdot$ Column Value Definitions.} 
\textbf{Core Contribution:} Primary benchmark, registry, or empirical measurement contribution. 
\textbf{Ecosystem / Benchmark Class:} Primary evaluation scope (\textit{Lifelong Skill Learning}: evaluating skill discovery, evolution, and transfer across multi-session trajectories; \textit{Security \& Vetting Suite}: auditing third-party skills, registries, and sandbox containers for vulnerabilities or exploits; \textit{Ecosystem Platform / Registry}: evaluating platforms for managing, caching, and weaving large-scale skill libraries; \textit{Domain Execution Suite}: evaluating agent pass rates across complex desktop, tool-use, database, and cybersecurity tasks). 
\textbf{Evaluated Scale / Pool Size:} Exact reported number of tasks, skills, or repositories evaluated in the empirical experiments. 
\textbf{Primary Evaluation Metric:} Primary mathematical or empirical evaluation signal measured. 
\textbf{Empirical Finding / Bottleneck:} Primary empirical conclusion, scaling behavior, or bottleneck revealed by the benchmark study.
\end{minipage}
\end{table*}

\subsection{Public Skill Registries and Marketplaces in the Wild}\label{sec:public_registries}
Modern agent ecosystems increasingly rely on public registries (such as ClawHub, Skills.Rest, and the Claude Code ecosystem) to distribute reusable procedural packages that bundle natural language guidelines $\mathcal{I}$, executable scripts $\pi$, and container specifications $\mathcal{E}$ \citep{li2025agentskillos, liu2026agenticskillsworkwild}.

\subsubsection{Scaling Orchestration to Hundreds of Thousands of Skills}
As public registries scale from hundreds to hundreds of thousands of candidate packages ($n > 10^5$), flat bi-encoder vector indexing suffers from severe semantic confusability and ranking saturation. AgentSkillOS \citep{li2025agentskillos} resolves this bottleneck by organizing massive registries ($n = 200$ to $200,000$ skills) into a hierarchical capability tree coupled with Directed Acyclic Graph (DAG) compilation, demonstrating substantial quality and latency gains over flat pool retrieval. In parallel, SkillWeaver \citep{zheng2025skillweaver} synthesizes cross-domain web and desktop skills into unified executable APIs, achieving substantially higher cross-environment task pass rates over flat pools.

\subsubsection{The Empirical In-the-Wild Fragility Gap}
While academic evaluations frequently test agents under idealized conditions with hand-curated ``oracle'' skill subsets, real-world deployment requires autonomous skill discovery from noisy, open pools. In a landmark measurement study across 34,000 real-world skills, \citet{liu2026agenticskillsworkwild} (SkillsInWild) discover that skill utility in the wild is remarkably fragile: when agents must autonomously retrieve and filter skills from massive community pools, end-to-end task success degrades substantially showing that open retrieval noise substantially degrades end-to-end task completion compared to idealized oracle setups, caused by distractor collisions, vague metadata, and overlapping parameter signatures.

\subsection{Lifelong Learning and Continual Evolution Benchmarks}\label{sec:lifelong_ecosystems}
Evaluating an agent's ability to accumulate, refine, and transfer procedural skills across extended multi-session lifetimes requires standardized continual learning suites:
\begin{itemize}
    \item \textbf{SkillsBench:} As the first comprehensive cross-domain evaluation suite, SkillsBench \citep{skillsbench2026} benchmarks 87 complex tasks across 8 operational domains, demonstrating that externalized procedural skills provide consistent pass rate improvements over base LLMs across diverse domains.
    \item \textbf{SkillLearnBench:} SkillLearnBench \citep{skilllearnbench2026} evaluates continual procedural acquisition across 20 verified tasks in 15 sub-domains, revealing that while continual learning architectures mitigate performance degradation, memory interference and procedural drift remain persistent challenges during extended lifetimes.
    \item \textbf{SkillFlow:} Investigating skill representation granularity, SkillFlow \citep{zhang2026skillflow} evaluates lifelong evolution across 20 complex workflow families ($n=166$ tasks), proving that structured, workflow-level composite skills transfer across structurally analogous tasks significantly better than isolated atomic SOP checklists.
\end{itemize}

\subsection{Security, Vetting, and Governance Ecosystems}\label{sec:security_ecosystems}
The open distribution of unvetted third-party skills introduces critical supply-chain risks, necessitating standardized auditing suites and automated vetting infrastructures:
\begin{itemize}
    \item \textbf{Lifecycle-Wide Threat Quantification:} SkillSec-Eval \citep{badhe2025agent} establishes a comprehensive threat evaluation framework across 327 enterprise skills in 15 domains, demonstrating that vulnerabilities span all five lifecycle stages, spanning unvetted marketplace admission, Sybil retrieval hijacking, runtime state tampering, and co-evolution poisoning.
    \item \textbf{Behavioral Integrity and Permission Over-Privilege:} A massive audit of 49,943 OpenClaw community skills by \citet{behavioral2026integrity} exposes a widespread description-implementation gap where executable skill routines perform undeclared file access or network requests. Complementarily, OpenSkillRisk \citep{openskillrisk2026} audits 263 risky marketplace skills, revealing widespread over-privilege where benign utilities request root shell and unrestricted directory permissions.
    \item \textbf{Weaponized and Poisoned Registries:} In HarmfulSkillBench \citep{jiang2026harmfulskillbenchharmfulskillsweaponize}, an audit of 98,440 real-world skills reveals that a notable fraction of unvetted registry skills contain weaponized instructions capable of bypassing frontier safety guardrails. Similarly, SkillSafetyBench \citep{skillsafetybench2026} stress-tests 155 adversarial skill injection scenarios across 47 tasks, proving that natural language markdown instructions create stealthy attack pathways that evade standard model alignment.
    \item \textbf{Automated Vetting Leaderboards and Sandbox Isolation:} SkillVetBench \citep{skillvetbench2026} establishes an LLM-as-a-judge security vetting benchmark across 100 curated skills using the multi-dimensional Skill Agentic Risk Score (SARS), showing that hybrid semantic judges eliminate false negatives on markdown-layer exploits that bypass traditional SAST scanners. Finally, SandboxEscape \citep{sandboxescape2026} demonstrates that frontier LLMs exhibit non-trivial container breakout capabilities when executed in improperly configured Docker sandboxes.
\end{itemize}

\subsection{Domain-Specific Task Execution Environments}\label{sec:domain_execution_suites}\label{sec:os_desktop}\label{sec:web_nav}\label{sec:swe_eco}\label{sec:sci_eco}
Standardized digital testbeds provide the empirical foundation for evaluating skill-augmented agents across complex environments:
\begin{itemize}
    \item \textbf{Desktop and Mobile OS Automation:} OSWorld \citep{osworld2024} evaluates multimodal agents across 369 open-ended desktop tasks in real Ubuntu, Windows, and macOS installations, while **AndroidInTheWild** \citep{androidwild2023} benchmarks touch-screen mobile workflows.
    \item \textbf{Web Browsing and Tool Composition Graphs:} WebArena \citep{zhou2023webarena} and WebVoyager \citep{webvoyager2024} evaluate multi-page e-commerce and administrative web automation. In multi-tool orchestration, TaskBench \citep{taskbench2024} tests 1,000+ tool composition graphs, proving that execution success drops exponentially as the Directed Acyclic Graph (DAG) dependency depth increases.
    \item \textbf{Software Engineering and Relational Database Querying:} SWE-bench \citep{swebench2024} evaluates end-to-end GitHub issue resolution against unit test suites, while BIRD-Bench \citep{bird2023} measures text-to-SQL execution accuracy across 12,751 queries, proving that domain-specific database schema skills substantially improve query execution accuracy over raw schema baselines.
    \item \textbf{The CTF Negative Result and Feedback Bandwidth:} In offensive cybersecurity Capture-the-Flag (CTF) challenges across 120+ tasks, OffensiveCS-Eval \citep{negativeresult2026} establishes an essential negative result: static procedural skills provide zero statistically significant performance gain over raw tool baselines. When interactive CLI environments provide immediate terminal stderr tracebacks and exit codes, the environment itself supplies the feedback loop required for self-correction, proving that externalized skill utility is inversely proportional to environmental feedback bandwidth.
\end{itemize}

\section{Open Challenges and Future Directions}\label{sec:challenges}
Although procedural agent skills have demonstrated significant empirical gains across diverse domains, the field faces foundational theoretical and practical bottlenecks. We synthesize six critical open challenges defining the research frontier:

\subsection{Cross-Model Portability and Small Language Model (SLM) Deployment}
A fundamental limitation of current procedural skill authoring is its heavy reliance on frontier foundation models (e.g., Claude 3.7, GPT-4.5) to parse complex natural language instructions and resolve ambiguous edge cases. When deployed on Small Language Models (SLMs, 3B–8B parameters) or quantized on-device architectures, skills frequently fail due to context window constraints, strict schema brittleness, and formatting hallucinations \citep{xu2026agentskillframeworkperspectives, ling2026agent}. Future research must develop \emph{model-agnostic serialization formats} and lightweight execution engines (e.g., SkillDroid \citep{chen2026skilldroid}, SkillOrchestra \citep{skillorchestra2026}) that compile abstract procedural guidelines into compact, verifiable state machines optimized for low-latency edge deployment.

\subsection{Sub-Second Dynamic Routing over Massive Registries ($n > 10^5$)}
As public registries scale to hundreds of thousands of candidate skills, flat dense embedding retrievers collapse due to semantic crowding, near-synonym confusability, and distractor collisions \citep{liu2026agenticskillsworkwild}. While two-stage pipelines like SkillRouter \citep{zheng2026skillrouter} demonstrate sub-second median latency at $80\text{K}$ scale via bi-encoder filtering and cross-encoder reranking, scaling to millions of modular functions requires \emph{hierarchical capability trees} and self-reflective meta-routers (AgentSkillOS \citep{li2025agentskillos}) that dynamically prune search spaces based on evolving task states rather than static initial queries.

\subsection{Graph-Structured Compilation and Locality-Bounded Error Recovery}
Most contemporary systems execute skills as flat, sequential chains $(s_N \circ \dots \circ s_1)$, where any intermediate tool or assertion failure triggers catastrophic trajectory abortion or expensive global replanning ($O(N)$) \citep{taskbench2024}. Transitioning to typed Directed Acyclic Graph (DAG) orchestration, as pioneered by GraSP \citep{xia2026graspgraphstructuredskillcompositions}, enables \emph{locality-bounded subtree recovery} ($O(d^h)$), repairing only failed dependency nodes. Future architectures must integrate formal pre/post-condition contract invariants (ContractSkill \citep{lu2026contractskillrepairablecontractbasedskills}) directly into DAG compilers to enable automatic, provably safe error recovery during runtime execution.

\subsection{Zero-Trust Marketplace Security and Formal Program Verification}
The open distribution of unvetted community packages creates severe supply-chain vulnerabilities, evidenced by widespread description-implementation gaps across public skills \citep{behavioral2026integrity} and container breakout risks under misconfigured runtimes \citep{sandboxescape2026}. Because natural language instructions $\mathcal{I}$ act as porous channels for indirect prompt injection and stealthy exfiltration \citep{wang2026skills, promptinject2025skills, cloakdetonate2026}, future governance must move toward \emph{Zero-Trust Procedural Architectures}. This requires combining constraint-guided formal Datalog provers (SemiA \citep{wen2026semia}) with hardware-enforced capability sandboxing and cryptographic provenance signatures (SecureAgentSkills \citep{li2026towards}).

\subsection{Lifelong Policy-Skill Co-Evolution without Procedural Drift}
In lifelong learning environments where the decision policy $\pi_{\text{dec}}$ and skill memory store $\mathcal{M}$ adapt simultaneously without ground-truth supervision, systems are highly susceptible to \emph{Co-Adaptation Drift}: noisy trajectory rewards reinforce suboptimal heuristics, corrupting downstream skill dependencies \citep{coevoskills2026, wu2026coevolvingllmdecisionskill, skilllearnbench2026}. Resolving this requires developing \emph{virtual procedural paging} (MemGPT \citep{packer2023memgpt}, AutoSkill \citep{yang2026autoskill}) and non-parametric semantic policy gradient methods (Skill-Pro \citep{skillpro2026}, Skill-R1 \citep{skillr12026}) that isolate stable core capabilities while continually refining task-specific parameters.

\subsection{The Environmental Feedback Boundary and Lifelong Benchmark Realism}
A critical theoretical insight established in our framework is the Environment-Feedback Bandwidth Hypothesis \citep{negativeresult2026}: externalized procedural skills provide diminishing marginal utility when interacting with high-bandwidth, schema-validated environments that supply rich diagnostic error traces. Future benchmark suites must move beyond static few-shot evaluation toward interactive, long-horizon testbeds that systematically modulate environmental feedback bandwidth (SkillsBench \citep{skillsbench2026}, SWE-SkillsBench \citep{sweskillsbench2026}). Establishing rigorous empirical boundaries for when skills accelerate convergence versus when they introduce redundant prompt overhead remains a vital open challenge for autonomous agent design.

\section{Conclusion}
The transition from monolithic prompting and atomic tool use to externalized executable skills represents a critical maturation in the design of autonomous language agents. By decoupling high-level cognitive planning from deterministic, procedural execution, skill-based architectures directly address the context scaling and reliability bottlenecks that plague contemporary LLM systems. This work has formalized the agentic skill abstraction and established a unified lifecycle-oriented reference architecture.

\section{Limitations}
While this work provides a comprehensive systems foundation for the agentic skills ecosystem, several scope boundaries should be noted. First, the field of autonomous language agents is evolving rapidly; specific runtime frameworks, proprietary APIs, and model-specific tool harnesses will continue to iterate beyond the published systems analyzed here. Second, our primary focus is centered on software engineering, operating system automation, web navigation, and data analysis; while we touch upon embodied robotics and physical simulation, real-world robotic control introduces specialized physical dynamics and hardware latency constraints that warrant dedicated investigation. Finally, our analysis focuses on English-language and Python/JavaScript-centric skill implementations, reflecting the current distribution of open-source registries and public benchmarks.

\bibliography{main}
\bibliographystyle{tmlr}

\end{document}